\documentclass[10pt,twocolumn,letterpaper]{article}

 \usepackage[pagenumbers]{cvpr} 

\definecolor{cvprblue}{rgb}{0.21,0.49,0.74}
\usepackage[pagebackref,breaklinks,colorlinks,allcolors=cvprblue]{hyperref}
\usepackage{caption} 
\usepackage{booktabs}
\usepackage{amsmath}
\usepackage{tabularx}
\usepackage{multicol}
\usepackage{multirow}
\usepackage{xcolor}
\usepackage{algorithm}
\usepackage{algpseudocode}
\usepackage{amsmath}
\usepackage{newfloat}
\usepackage{listings}
\def\paperID{1768} 
\def\confName{CVPR}
\def\confYear{2025}

\title{	
Seeing Clearly by Layer Two: Enhancing Attention Heads to Alleviate Hallucination in LVLMs}

\author{
Xiaofeng Zhang*\textsuperscript{\rm 1},
Yihao Quan*\textsuperscript{\rm 3},
Chaochen Gu$\dagger$\textsuperscript{\rm 1},
Chen Shen\textsuperscript{\rm 2},
Xiaosong Yuan\textsuperscript{\rm 2},
Shaotian Yan\textsuperscript{\rm 2},\\
Hao Cheng\textsuperscript{\rm 1},
Kaijie Wu\textsuperscript{\rm 1},
Jieping Ye\textsuperscript{\rm 2}\\
\textsuperscript{\rm 1}Shanghai Jiao Tong University \\
\textsuperscript{\rm 2}Alibaba Group \quad \
\textsuperscript{\rm 3}Beijing Jiaotong University \\
{\tt\small \{framebreak@\}sjtu.edu.cn}\quad
{\tt\small \{21711103@\}bjtu.edu.cn} }

\begin{document}

\maketitle
\def\thefootnote{*}\footnotetext{These authors contributed equally to this work}
\begin{abstract}
The hallucination problem in multimodal large language models (MLLMs) remains a common issue. Although image tokens occupy a majority of the input sequence of MLLMs, there is limited research to explore the relationship between image tokens and hallucinations. In this paper, we analyze the distribution of attention scores for image tokens across each layer and head of the model, revealing an intriguing and common phenomenon: most hallucinations are closely linked to the pattern of attention sinks in the self-attention matrix of image tokens, where shallow layers exhibit dense attention sinks and deeper layers show sparse attention sinks. We further analyze the attention heads of different layers and find that heads with high-density attention sink in the image part play a positive role in alleviating hallucinations. In this paper, we propose a training-free method named \textcolor{red}{\textbf{E}}nhancing \textcolor{red}{\textbf{A}}ttention \textcolor{red}{\textbf{H}}eads (EAH), an approach designed to enhance the convergence of image tokens attention sinks in the shallow layers. EAH identifies the attention head that shows the vision sink in a shallow layer and extracts its attention matrix. This attention map is then broadcast to other heads in the layer, thereby strengthening the layer to pay more attention to the image itself. With extensive experiments, EAH shows significant hallucination-mitigating performance on different MLLMs and metrics, proving its effectiveness and generality.

\end{abstract}

\section{Introduction}
\begin{figure}[t]
\centerline{\includegraphics[scale=0.62]{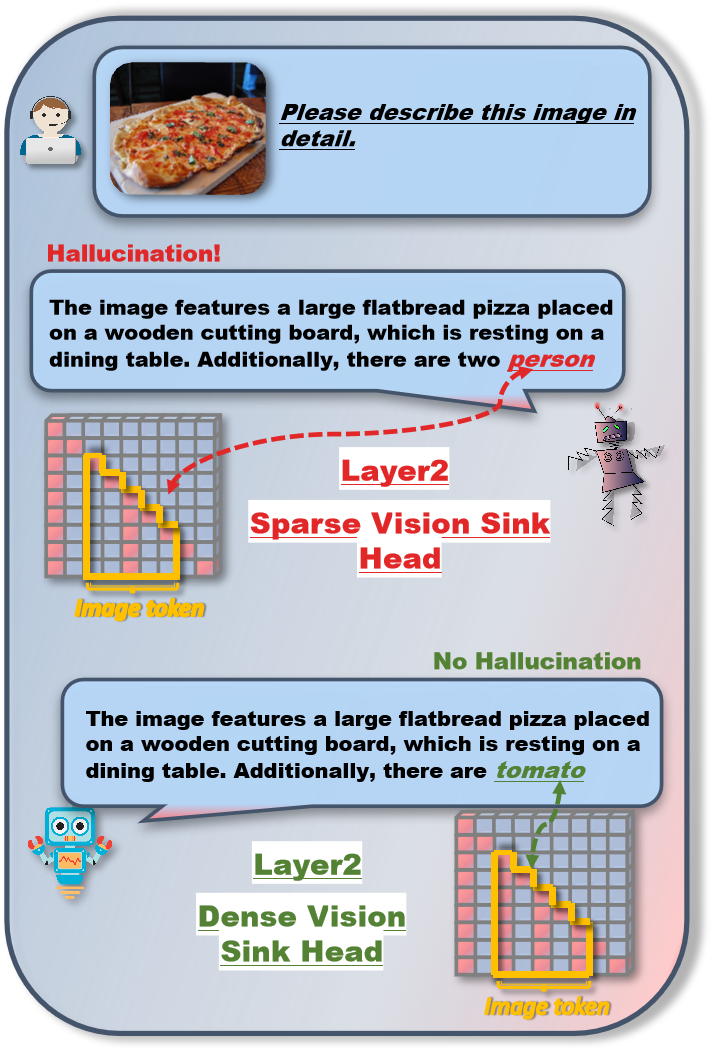}}
\caption{We found a common phenomenon through the attention map: In the range of image token, the attention head of shallow Sparse attention sink is prone to hallucination, while the attention head of Dense attention sink is much less likely to hallucinate.}
\label{intro}
\end{figure}

Multimodal large language models (MLLMs) \cite{GPT4V,LLaVA,bai2023qwen,minigpt4,instructblip,LLAMAVID,team2023gemini,gong2023multimodal,zhang2023video} have made significant strides in cross-modal tasks, especially in handling both text and image modalities. However, hallucinations remain a persistent challenge, particularly in tasks such as Visual Question Answering (VQA) or image captioning. Current methods for addressing hallucinations often involve changing decoding strategies, incorporating external knowledge bases, or retraining models with additional data \cite{li2023fine,liu2023aligning,park2024mitigating}. These approaches, however, often require significant resources and time.

Recent research into attention sink has offered new insights into hallucinations. The concept of attention sink as an information flow is introduced in ``Label Words are Anchors" \cite{label-words}, which shows how information flow often converges on a specific user token in large language models (LLMs). OPERA \cite{opera} further explores the connection between attention sink in user tokens and output tokens in MLLMs. It observes that when a token has a high attention weight across subsequent tokens, this over-reliance on the token can lead to hallucinations in the model's outputs. Although these methods clarify the relationship between attention sink, user tokens, and output tokens, the relationship between attention sink, image tokens, and hallucination remains unclear. It's important to note that MLLM's output tokens are generated by the decoder based on logits, whereas input tokens, which constitute most of the input sequence, are more likely to directly reflect the MLLM's internal mechanisms.

\begin{figure}[t]
\centerline{\includegraphics[scale=0.32]{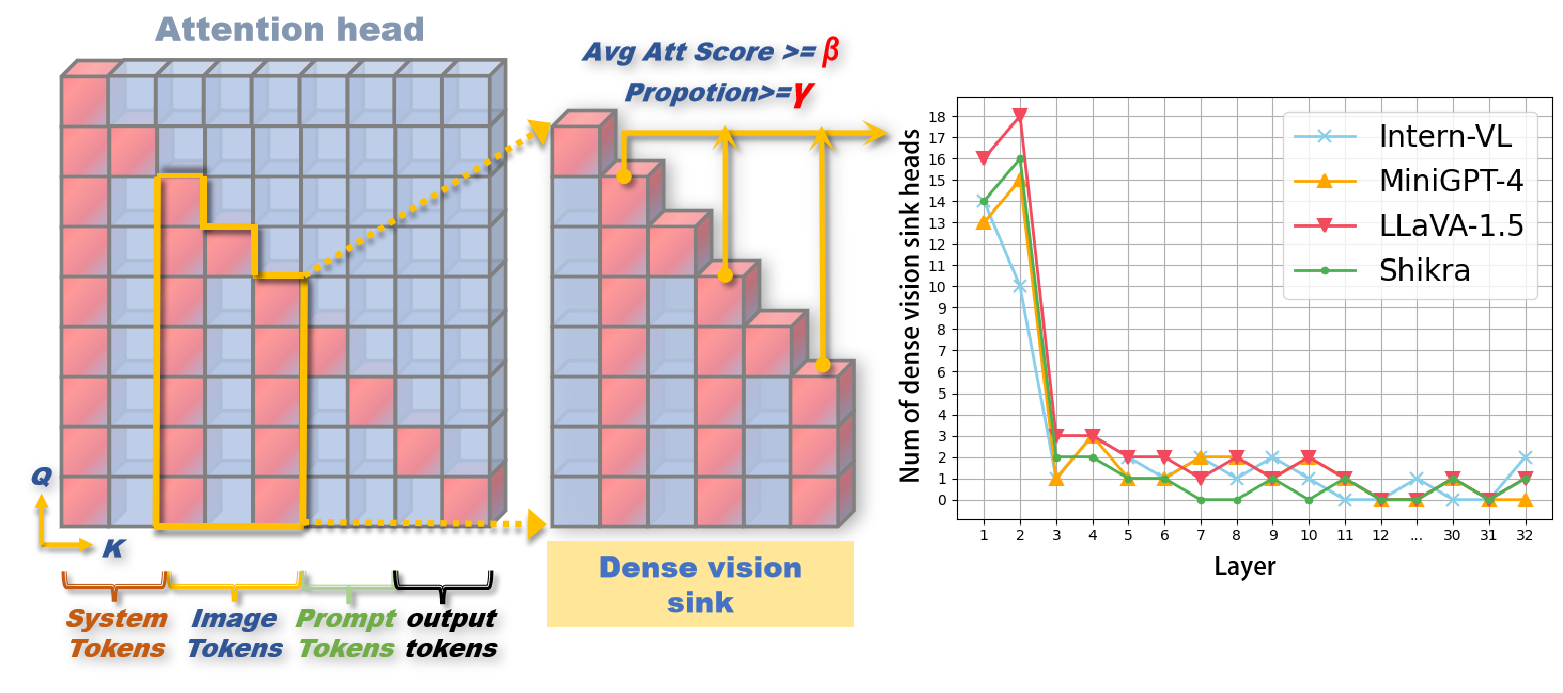}}
\caption{Definition of dense vision sink head and its layer-wise distribution. In this case, $\beta$ = 0.0015, $\gamma$ = 15\%}
\label{introduction2}
\end{figure}

\begin{figure}[h]
\centerline{\includegraphics[scale=0.32]{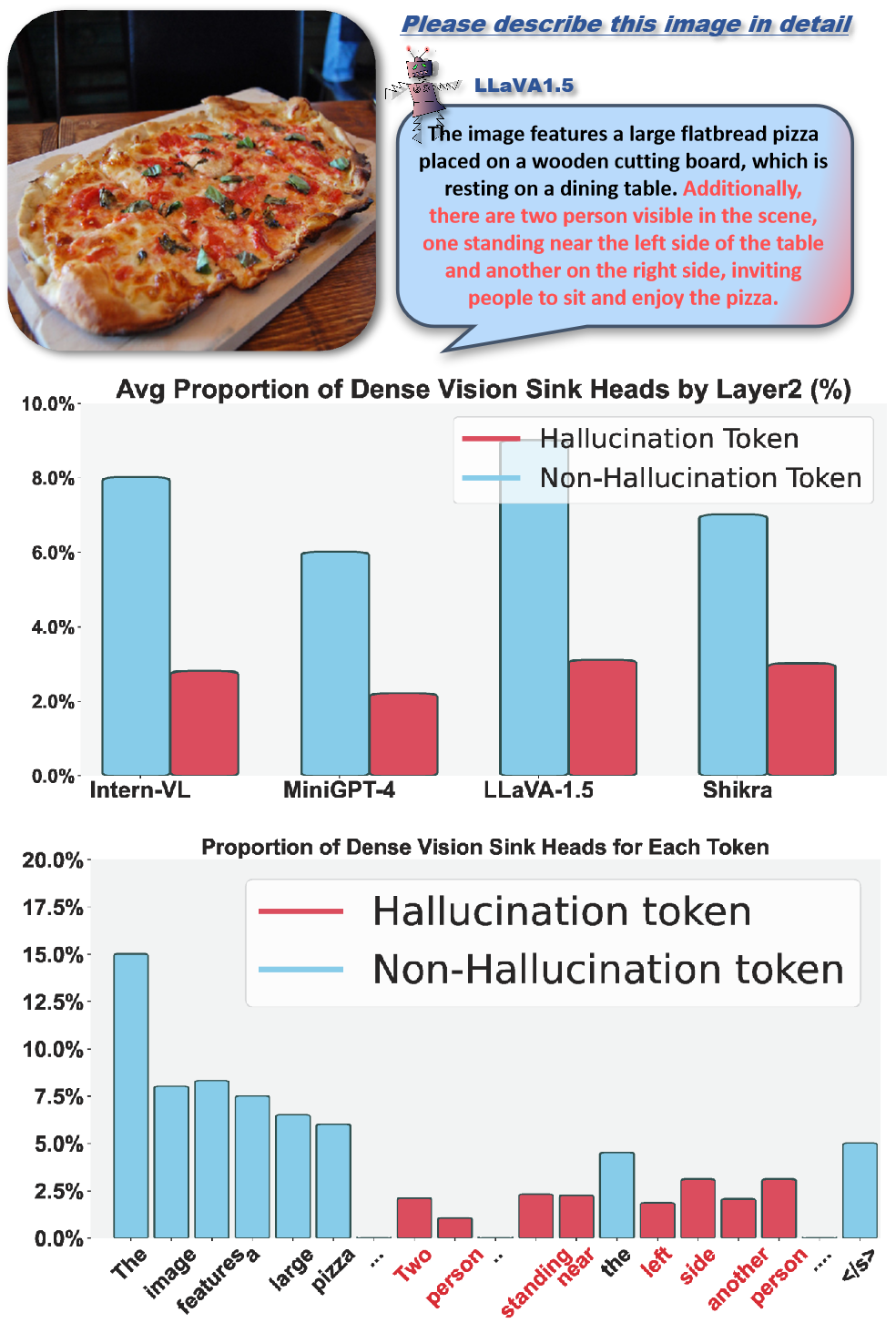}}
\caption{Relationship between text tokens and the average proportion of dense vision sink heads within a single layer by layer2, analyzed across 5,000 randomly selected MSCOCO images using LLaVA1.5-7B.}
\label{motivation1}
\end{figure}

\textbf{Most dense vision sink heads occur by layer2:} As previously mentioned by FastV \cite{fastv}, the information flow of image tokens is primarily concentrated in the first and second layers. Building on this, we conduct experiments on the shallow layers of several models, including LLaVA1.5 \cite{LLaVA}, Minigpt4 \cite{minigpt4}, MiniGemini \cite{minigemini}, and Intern-VL \cite{internvl}. As shown in Fig. \ref{introduction2}, we calculate the average number of dense vision sink heads across these layers to further investigate the distribution of attention sinks across different layers.

We define $h_{i,j}$ as the attention-map of a head, a ``dense vision sink head" as a head \( (i, j) \) in which the proportion \( \alpha^{i,j} \) of columns in the attention map that meet the vision sink condition exceeds a threshold \( \gamma \). Specifically, we define \( \alpha^{i,j} \) as:

\begin{equation}
\text{vision sink} = \frac{\sum_{x=k}^{r} h_{i,j}[x][y] \cdot M}{r-k} > \beta, \quad k \in [36, 611],
\end{equation}

\begin{equation}
\alpha^{i,j} = \frac{\text{Num(vision sinks)}}{576},
\end{equation}

A head is classified as a ``dense vision sink head'' if:

\begin{equation}
\alpha^{i,j} \geq \gamma.
\end{equation}

Observations show that most vision attention sinks occur by layer 2.





\textbf{Fewer dense vision sink heads lead to hallucination output:} 
\begin{equation}
p = \frac{\text{Num(dense vision sink heads)}}{32}
\end{equation}
This proportion \( p \) quantifies how many of the total 32 heads are classified as "dense vision sink heads," meaning they have a high proportion of columns that meet the vision sink condition within the image token range.
We conducted the image captioning task on 5,000 randomly selected MSCOCO images using LLaVA1.5-7B. When the model generates a new token, we first determine whether it is a hallucination token. Then, we backtrack to layer 2 and analyze the model at the granularity of attention heads. We calculate the proportion of dense vision sink heads in layer 2 relative to the total number of heads (e.g., 32 heads in total). This analysis is repeated for layer 1, and the average across layers 1 and 2 is then computed. As shown in Fig. \ref{motivation1}, we observe that non-hallucination tokens typically activate a larger number of dense vision sink heads, whereas hallucination tokens are generally associated with only a few dense vision sink heads, with the majority of heads being sparse. Through our analysis of different models, such as LLaVA1.5 \cite{LLaVA}, Minigpt4 \cite{minigpt4}, MiniGemini \cite{minigemini} and Intern-VL \cite{internvl}, it appears that fewer dense vision sinks heads lead to more probable hallucination output. 
\begin{figure}[t]
\centerline{\includegraphics[scale=0.16]{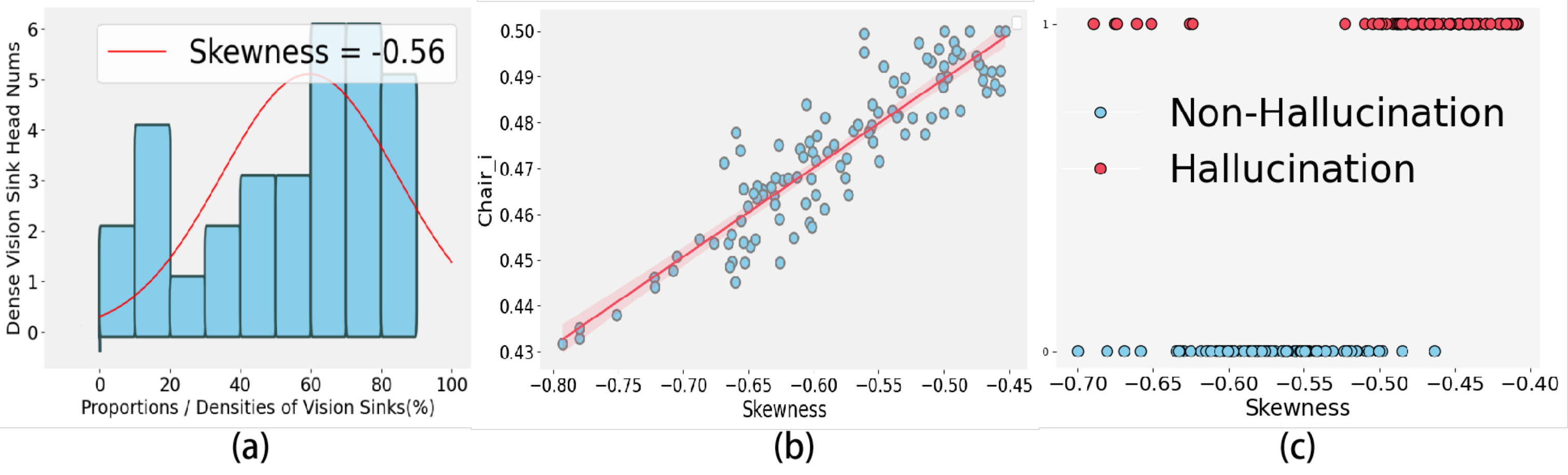}}
\caption{(a) A example of distribution of dense vision head  and the corresponding proportions/densities of vision sinks within these heads when model output hallucination token; (b) Relationship between the average skewness and CHAIR$_{I}$ on 150 randomly selected MSCOCO images, using LLaVA1.5-7B for captioning; (c) Comparative skewness scatter plot for Hallucination and Non-Hallucination classification on 150 randomly selected MSCOCO images, using LLaVA1.5-7B for VQA.}
\label{motivation4}
\end{figure}

\textbf{Lower density of vision sinks and fewer vision sink heads lead to a higher probability of hallucinations:} However, the average number of dense vision sink heads across shallow layers does not reveal the individual contributions of each dense vision head, some of which may be negative while others are positive for hallucination. As noted by ITI \cite{ITI}, in current large language models (LLMs) using transformer architecture, only a subset of attention heads plays a more significant role. Effectively optimizing these heads and leveraging them will likely lead to substantial improvements in model efficiency and overall performance. In this case, we conducted a more detailed view for each head, as shown in Figure \ref{motivation4} (a), the sink densities within different vision sink heads vary across the shallow layers (layer1-layer2), with an overall negatively skewed distribution. As shown in Figure \ref{motivation4} (b), for the image captioning task, the average skewness of the distribution of dense vision sink head and its corresponding vision sink densities in layer1 and layer2 is recorded each time a token is output. Once the output token is completed, the CHAIR$_{I}$ for the entire output is calculated, and the average skewness for all tokens in layer1 and layer2 is obtained. As shown in Figure \ref{motivation4} (c), for the VQA task (with only a single output token), the average skewness of the distribution of vision sink head and its corresponding vision sink densities in layer1 and layer2 is directly recorded for the answer token. It is observed that, regardless of the task (image captioning or VQA), a lower skewness coefficient correlates with a lower hallucination rate. In other words, a higher density of vision sinks within a dense vision sink head and a larger number of vision sink heads lead to a lower probability of hallucination.

These observations highlight the critical role of attention head and vision sink distribution in understanding the attention sink phenomenon, particularly as it relates to alleviating hallucination issues in MLLMs. When the vision sink is sparse, visual tokens concentrate too heavily on specific elements, leading to reduced attention to other parts of the image. Conversely, a dense vision sink helps maintain a global perspective, preventing the model from narrowing its focus too much and minimizing information loss. Our goal is to ensure the model maintains a high-density vision sink within shallow layers. To achieve this, we design a training-free method called \textbf{E}nhancing \textbf{A}ttention \textbf{H}eads (EAH). This plug-and-play approach focuses on each attention head in the early layers, systematically identifying the head with the densest vision sinks. It then broadcasts this attention distribution across the layer, aligning the layer's attention and the head's vision sink distribution with that of the selected head.


We conduct extensive evaluations, focusing specifically on hallucination issues, and test mainstream MLLMs to validate the effectiveness of EAH in reducing hallucinations across various model architectures. Our results demonstrate that EAH is a highly effective plug-and-play solution for mitigating hallucinations across various MLLMs. Specifically, our contributions can be summarized as follows:

\begin{itemize}
\item 
This paper investigates how information flow relates to hallucinations in MLLMs. Our analysis reveals a consistent pattern where denser vision sinks and a larger number of vision sink heads in the shallow layers are associated with fewer hallucinations.
\item We propose a plug-and-play training-free method called \textbf{E}nhancing \textbf{A}ttention \textbf{H}ead, which alleviates hallucinations by finding the head with the densest vision sink and broadcasting it to other heads.
\item Experiments on multiple models validate the plug-and-play convenience and strong generalization of this method.
\end{itemize}


\section{Related Work}

\subsection{Attention Sink and Information Flow}
While the mechanisms of large language models (LLMs) and multimodal large language models (MLLMs) remain complex and not fully understood, several approaches focusing on information flow and attention sink patterns provide valuable insights into their operation and offer potential solutions to issues such as hallucinations and inefficiencies.

StreamingLLM \cite{efficient-attention-sink} first introduces the concept of attention sink. The authors observe an intriguing phenomenon: initial tokens, while seemingly less important for the overall content generation, consistently receive high attention scores. This is visualized in the attention map as columns with notably high attention scores, which is counterintuitive. Furthermore, because of the autoregressive nature of generative models, these initial tokens continue to receive more attention from subsequent tokens, amplifying their impact on the generation process. To address this, StreamingLLM leverages these attention-sink tokens during the pre-training phase to enhance the model's performance. 

In a similar vein, Label Words \cite{label-words} focuses on information flow within the model, identifying anchor tokens through saliency scores (attention score × gradient) as key to in-context learning (ICL). In Zero-shot ICL and Chain-of-Thought (CoT) tasks, these anchors align with the prompt, while in few-shot tasks, they converge on the final token, highlighting their role in effective learning and decision-making.

Building on this idea, ACT \cite{attention-sink} provides a deeper exploration of attention sinks in LLMs. By analyzing attention maps across various tasks and inputs, the study finds that not all attention sinks are beneficial for model accuracy. To mitigate this, ACT introduces an adaptive method for optimizing attention distributions during inference, aiming to improve model performance by selectively enhancing the most relevant attention patterns.

In the context of multimodal large language models (MLLMs), OPERA \cite{opera} introduces a novel perspective by linking the causes of hallucinations with attention sinks. This approach provides new insights into the interpretability of MLLMs. OPERA reveals that during the inference phase, the generation of key tokens such as '-', '?', or tokens that summarize previous ones can lead the model to produce hallucinated content. To address this issue, OPERA imposes penalty constraints on the attention scores of these summarization tokens. In light of this, DOPRA \cite{DOPRA} addresses the over-reliance by improving the strategy of weighted overlay penalties and redistribution in specific layers.

\begin{figure}[h]
\centerline{\includegraphics[scale=0.35]{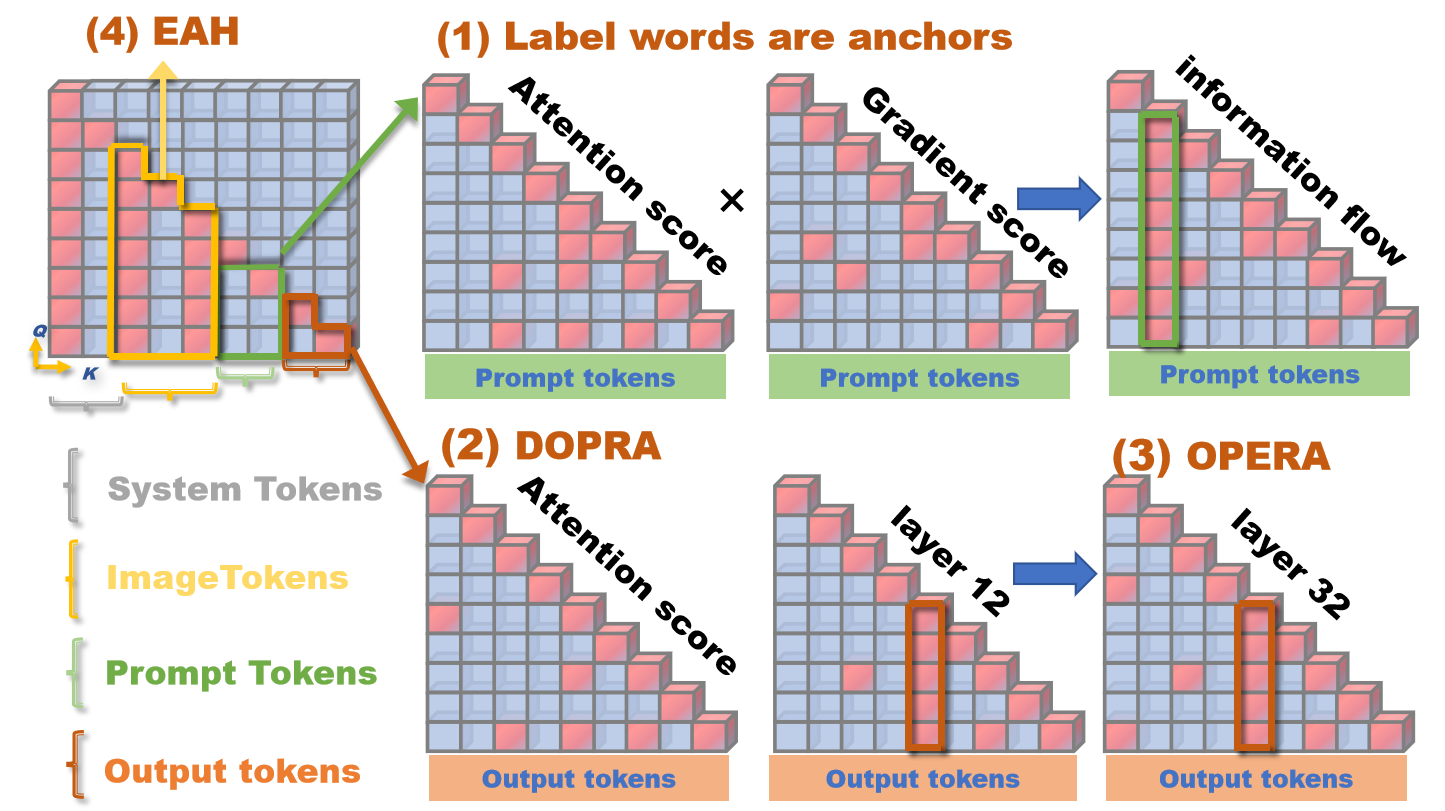}}
\caption{Differences in analysis perspectives between Label words \cite{label-words}, DOPRA \cite{DOPRA}, OPERA \cite{opera} and our method.}
\label{label-words}
\end{figure}

\textbf{Difference between These Methods:} 
The difference between EAH with Label-Words \cite{label-words}, OPERA \cite{opera} and DOPRA \cite{DOPRA} is shown in Fig. \ref{label-words}. The analysis method in Label Words focuses on prompt tokens, identifying anchor tokens based on saliency scores, which are the product of attention scores and gradient scores. OPERA \cite{opera} and DOPRA \cite{DOPRA} aim to reduce attention sinks by altering the decoding strategies for answer tokens. Our method, Enhancing Attention Head (EAH), enhances the convergence of image tokens' attention sinks in the shallow layers.

\begin{figure*}[t]
\centerline{\includegraphics[scale=0.5]{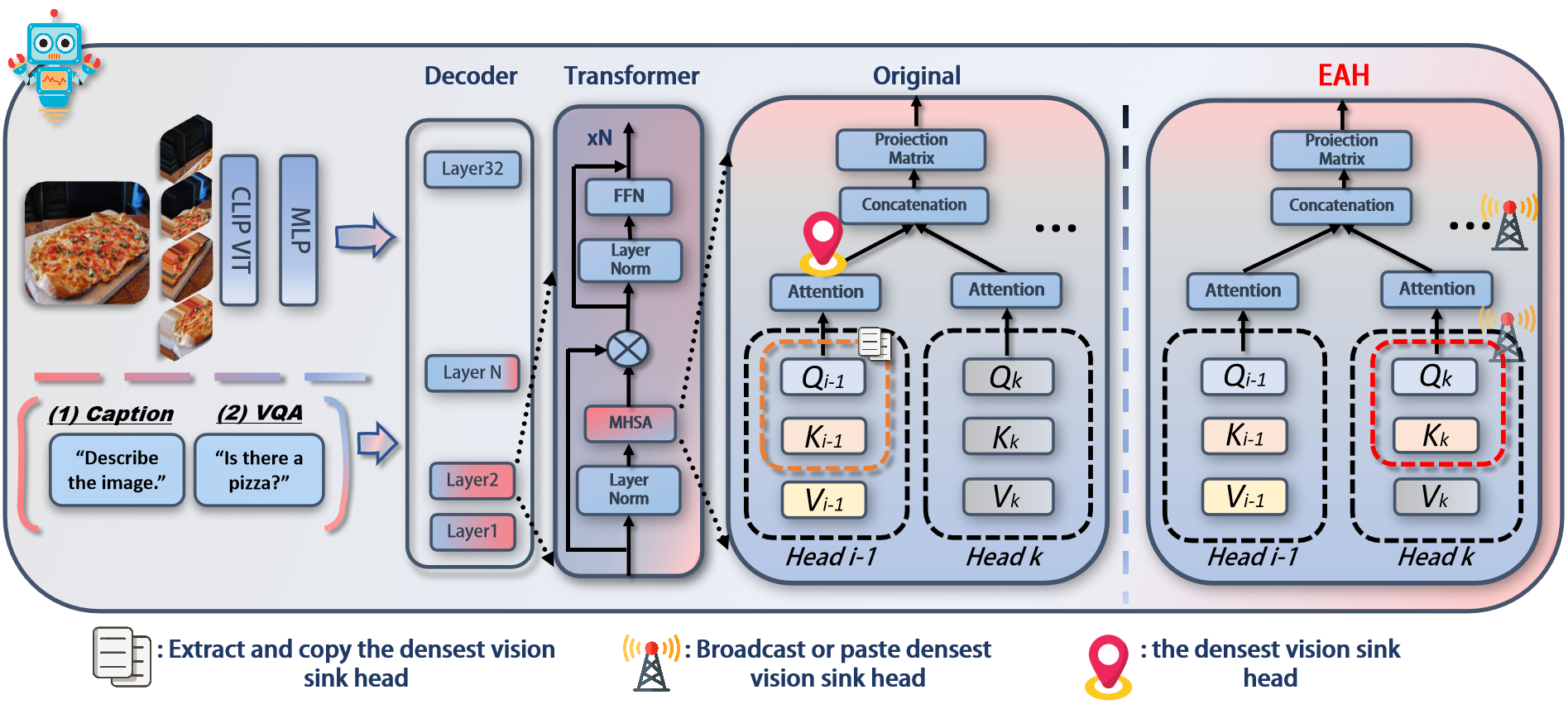}}
\caption{The structure of Enhancing Attention Head.}
\label{EAH}
\end{figure*}

\section{ Method}

\subsection{Relationship between Vision Sink and Hallucinations}
Popular VLMs, such as LLaVA-1.5 \cite{LLaVA}, Minigemini \cite{minigemini}, InstructBLIP \cite{instructblip}, Shikra \cite{shikra}, MiniGPT-4 \cite{minigpt4}, Qwen-VL \cite{bai2023qwen}, and InternVL \cite{internvl}, consistently exhibit a notable pattern: vision sinks are densely concentrated within the first and second layers, gradually becoming more sparse in deeper layers. As illustrated in Fig. \ref{introduction2}, Fig. \ref{motivation1}, and Fig. \ref{motivation4}, we conclude that a lower density of vision sinks and fewer vision sink heads correlate with an increased likelihood of hallucinations in model outputs. We hypothesize that maintaining dense attention sinks in shallow layers may help alleviate hallucinations, as concentrated attention in the early layers enhances the transfer of image information to subsequent layers. Therefore, a practical method is proposed to reduce hallucinations by ensuring a dense vision sink of attention heads by layer1 and layer2. Please refer to the \textbf{supplementary material} for more attention-map visualization results of different LVLMs.

\subsection{Vision Sink}
\textbf{Definition of Mask Matrix $M$.} To ignore diagonal elements in the attention map during calculations, we define a mask matrix $M$ as follows:

\begin{equation}
M \gets \operatorname{eye}(r, c) - \operatorname{diag}(1),
\end{equation}

where $\operatorname{eye}(r, c)$ generates an identity matrix of size $(r, c)$, and we set the diagonal elements to zero. 

\textbf{Definition of Vision Sink.} Let $h_{i,j}$ represent the attention map of the $j$-th head at the $i$-th layer, with $h_{i,j}[x][y]$ being the element at row $x$ and column $y$. We define a ``vision sink" as the column in the attention map within the image token range (e.g., $k \in [36, 611]$), where the average attention score of one element of the column within the image token range exceeds a threshold $\beta$.

For each column $y$, the "\textit{vision sink}" condition is defined as:

\begin{equation}
\textit{vision sink} = \frac{\sum_{x=k}^r h_{i,j}[x][y] \cdot M}{r - k} > \beta,
\end{equation}
where $k \in [36, 611]$.

\textbf{Definition of Dense Vision Sink Head:} For a head $(i, j)$, we calculate the proportion of columns that meet the vision sink condition, denoted as $\alpha^{i, j}$. If this proportion of vision sinks within the range of image tokens (e.g., 576) exceeds a preset threshold $\gamma$, we classify the head as a "dense vision sink head." which is defined as follows:

\begin{equation}
\alpha^{i, j} = \frac{\text{Num(vision sink)}}{576} \geq \gamma.
\end{equation}



\subsection{ Enhancing Attention Head}

As mentioned above, we introduce a training-free, plug-and-play method called \textbf{E}nhancing \textbf{A}ttention \textbf{H}ead (EAH) to keep attention heads densely concentrated in the early layers. This approach identifies the attention head with the most dense attention sinks and broadcasts its attention map across other heads. This is to reinforce the attention pattern of a particular head or to broadcast the attention pattern under certain specific conditions (e.g. when a predefined threshold is exceeded).

The algorithmic process is shown in Algorithm \ref{alg:atten_process_cal}, let $A$ be a 4D tensor, where $A[i][j]$ denotes the attention matrix of the $j_{th}$ head of the $i_{th}$ layer. Let $\beta$ be the threshold, image-token-start-index be $s$, image-token-end-index be $e$ and $M$ be a mask matrix of the same size as $h_{i,j}$. We define $h_{i,j}$ as the attention-map of a head:




\textbf{Initialization.} Set the threshold \(\beta\) and initialize the variables: \(k\) as a randomly selected index within the range \([s, e]\), where \(s = 36\) is the starting index of the image tokens and \(e = 611\) is the end index; also initialize \(n = 0\) and $H = [\quad]$ (an empty list).

\textbf{Iteration over Heads.} Select a specific attention layer $i$ ($i \in \{0, 1, 2\}$), iterate on each head $j$ ($j \in [0, 31]$).

\textbf{Calculate Vision Sinks.} For each column $y$ in \( h_{i,j} \), calculate whether its column is a vision sink based on:
     \begin{equation}
     \text{id}_{i,j} = \{(x, y) \mid \frac{\sum_{r=k}^{r} h_{i,j}[x, y] \cdot M}{r - k} > \beta\},
     \end{equation} where \( \text{id}_{i,j} \) stores the indices \((x, y)\) where the average attention score exceeds \( \beta \).

\textbf{Store Count of Vision Sinks for each Attention Head.} Compute the count of marked indices for each head:
     \begin{equation}
     C_{i,j} = \text{count}(\text{id}_{i,j}),
     \end{equation}
     then append \( (C_{i,j}, j) \) to \( H \):
     \begin{equation}
     H = H \cup \{(C_{i,j}, j)\}.
     \end{equation}

\textbf{Update Head Index \( n \) with Maximum Vision Sinks.} Find the index \( n \) of the head with the maximum count \( C_{i,j} \) in \( H \):
     \begin{equation}
     n = \arg\max_{(C_{i,j}, j) \in H} C_{i,j}.
     \end{equation}This step dynamically updates \( n \) to track the head with the most vision sinks across the layer.

\textbf{Enhance Attention Heads across the Layer.} For each layer \( i \), set the matrix of head \( j \) with the head with the highest number of vision sinks to be the \( n \)-th position in \( A[i] \):
     \begin{equation} \text{for } j = 0, 1, \dots, 31: \quad A[i][j] = A[i][n].
     \end{equation}

\begin{algorithm}[t]
\small 
\caption{Attention Process Calculation, $nonzero()$ represents an operation with a non-zero index, and $\sum$ represents a sum, $h_{i,j}$ represents the $i$ layer, $j$ head, $k$ represents the index of token. $x$ and $y$ represent the rows and columns of $h_{i,j}$, respectively. }
\label{alg:atten_process_cal}
\begin{algorithmic}[1]
\Procedure{atten\_process\_cal}{$A$}
    \State Step 1: 
    $threshold \gets \beta$, $n \gets 0$\, $H \gets [ \quad ]$\
    \State Step 2: Loop over heads in a specific layer $i$, $i \in \{0, 1, 2\}$:\

        \For{$j \in \{0, 1, 2, \dots, 31\}$}:
        \State $h_{i,j} \gets A[i][j]$
        \State $s \gets 36$, $e \gets 611$
        
    \State Step 3: Calculate significant token indices:
        \State \qquad $M \gets \left( \text{eye}(r, c) - \operatorname{diag}(1)\right )$ 
        \State \qquad  $ids \gets \text{nonzero}\left( \frac{\sum_{x=k}^r  h_{i,j}[x,y] \cdot \operatorname{M}}{r-k} > \beta \right)$
    \State Step 4: Store the count of vision sinks and its corresponding head index in $H$: 
        \State \qquad $C_{i,j} \gets \text{count}(\text{id}_{i,j})$, $H \gets H \cup \{(C_{i,j}, j)\}$

        \State Step 5:  Update head index $n$ with maximum vision sinks: 
        \State   \qquad $n = \arg\max_{(C_{i,j}, j) \in H} C_{i,j}$
        \EndFor
        \For{$j \in [0, 31]$}
            \State $A[i][j] \gets A[i][n]$  
        \EndFor   
    \State \Return updated $A$
\EndProcedure
\end{algorithmic}
\end{algorithm}

\section{Experiment}
\subsection{Experimental Setup and Dataset}
\noindent\textbf{POPE evaluation on hallucinations.} POPE \cite{pope} is a dataset designed to evaluate hallucinations at the object level in question-answering tasks. It consists of a series of true/false questions about images, such as \texttt{"Is there a dog in the image?"}. Given a dataset of images and their corresponding object annotations, POPE constructs triples that include an image, a question, and a response.

\noindent\textbf{CHAIR evaluation on hallucinations.} The Caption Hallucination Assessment with Image Relevance (CHAIR) \cite{chair} metric is designed to evaluate hallucinations in object-level image captioning tasks. It includes two key dimensions: CHAIR$_{I}$ measures the proportion of objects mentioned in the caption that are not present in the ground-truth image, while CHAIR$_{S}$ assesses the proportion of captions that contain hallucinations.
\begin{equation}
    \scriptsize C_S = \frac{|\{\text{hallucinated objects}\}|}{|\{\text{all mentioned objects}\}|}, C_I = \frac{|\{\text{captions w/ hallucinated objects}\}|}{|\{\text{all captions}\}|},
\nonumber
\end{equation}
\noindent\textbf{Evaluation on general
vision-language tasks.} We also evaluate EAH on a variety of visual-verbal benchmarks, including general visual-linguistic tasks and vision-centered tasks such as MME-Bench \cite{mme}, MM-Vet \cite{mm-vet}, VizWiz \cite{vizwiz}, VQA$v2$ \cite{vqav2}, SEED \cite{seed}, GQA \cite{gqa}, and Blink \cite{blink}. SEED contains 19k multiple-choice questions with human annotations and covers 12 evaluation dimensions, including image and video data. MME-Bench also examines the general perception capabilities of LVLMs using a wide range of tasks. Additionally, we compare against VizWiz and GQA, which assess specific perception capabilities, such as knowledge and relations.
\subsection{Evaluation Results of EAH on 
Hallucination Benchmarks}
It is shown in Table \ref{compare-table}, that the methods to mitigate hallucinations can be broadly classified into four groups. The first group includes OPERA \cite{opera}, DOPRA \cite{DOPRA}, VCD \cite{vcd}, HACL \cite{halc} and AGLA \cite{agla}, which address hallucinations by altering the decoding process. The second group, represented by LESS is more \cite{less}, modifies the logits of the end-of-sequence (EOS) symbol to control its positioning, allowing the model to terminate earlier and reduce hallucinations. The third group is CCA-LLaVA \cite{causal-attention}, which explores the weakened information flow between visual and instruction tokens caused by the long-term decay in rotational position encoding (RoPE). They propose Concentric Causal Attention (CCA) to alleviate the object hallucination problem by reorganizing the positions of visual tokens and correcting the causal mask. The fourth group includes ITI \cite{ITI} and EAH, which aim to adjust attention heads to enhance the truthfulness of the model’s output during inference. Among these methods, EAH demonstrates competitive performance and achieves notable results. Among these methods, EAH demonstrates competitive performance, ranking among the top three alongside FastV \cite{fastv} and LESS \cite{less}.

\begin{table}
  \centering
  \scalebox{0.83}{
  \begin{tabular}{c|c|cccc}
    \toprule
    \multirow{2}{*}{Method} & \multicolumn{1}{c|}{POPE \cite{pope} }  & \multicolumn{4}{c}{CHAIR \cite{chair}} \\
    \cmidrule{3-6}
    & F1 score$\uparrow$  & \textit{C\( _{S} \)} $\downarrow$ &\textit{ C\( _{I} \)} $\downarrow$ & \textit{Recall} & \textit{Avg. Len} \\
    \midrule
    Greedy Search        & 85.7 & 47.0   & 13.8  & 76.6 & 94.2 \\
    Beam Search    & 84.9 & 51.0   & 15.2  & 75.2 & 102.2  \\
    DoLa  \cite{dola}         & 80.2 & 57.0   & 15.2  & 78.2 & 97.5 \\
    ITI \cite{ITI}   & 83.7 & 48.2   & 13.9  & 78.3 & 98.6 \\
    VCD   \cite{vcd}         & 83.2 & 51.0   & 14.9  & 77.2 & 101.9  \\
    AGLA \cite{agla} &   84.6 & 43.0   & 14.1 & 78.9 & 98.8 \\
    OPERA \cite{opera}         & 85.2 & 47.0   & 14.6  & 78.5 & 95.3 \\
    DOPRA \cite{DOPRA} & 85.6 & 46.3   & 13.8  & 78.2 & 96.1 \\
    HALC \cite{halc} & 83.9 & 50.2 & 12.4 & 78.4 & 97.2 \\
    FastV \cite{fastv} & 81.3 & \underline{39.4} & \underline{11.3} & 69.5 & 90.0 \\
    Less is more \cite{less} & \textbf{86.0} & 40.2   & 12.3  & 75.7 & 79.7 \\
    CCA-LLaVA \cite{causal-attention} & 85.5 & 43.0   & 11.5  & 80.4 & 96.6 \\
    EAH  & \underline{85.7} & \textbf{36.4} & \textbf{9.9} & 74.9 & 97.7 \\
    \bottomrule
  \end{tabular}}
    \caption{Compare results of EAH with other SOTA methods on POPE and CHAIR datasets. We report the average $F1-score$ computed on random, popular, and adversarial splits of POPE (baseline: LLaVA-1.5-7B), max-tokens=512, $Layer=2$, $\beta=0.002$. The best performances within each setting are \textbf{bolded}.}
      \label{compare-table}
\end{table}

Table \ref{compare-table} presents our validation of EAH using POPE and CHAIR on the LLaVA1.5-7B model \cite{LLaVA}, along with the mean $F1$ scores. We also provide a comparison of EAH with several other methods, including Dola \cite{dola}, ITI \cite{ITI}, VCD \cite{vcd}, OPERA \cite{opera}, DOPRA \cite{DOPRA}, AGLA \cite{agla}, Halc \cite{halc}, Less is More \cite{less}, FastV \cite{fastv} and CCA-LLaVA \cite{causal-attention}. FastV's token-cutting approach appears to be less effective for VQA datasets like POPE \cite{pope}. Although LLaVA’s training alignment may lead to some redundant image tokens, removing these tokens based on attention scores could inadvertently discard important local information, thus diminishing performance on the POPE dataset.

For captioning tasks, such as those on the CHAIR dataset, FastV \cite{fastv} and Less \cite{less} achieve second and third place, respectively. Their success lies in their core approach of reducing token generation, which helps mitigate hallucination. However, FastV \cite{fastv} encounters challenges with recall when the average token length is short, as it struggles to balance hallucination reduction with maintaining output length. While Less \cite{less} achieves state-of-the-art performance on the POPE dataset, its shorter average token length (79.7) results in performance that is less comparable with other methods. These findings underscore the effectiveness of EAH, which achieves the best performance by focusing solely on adjusting attention heads without compromising the integrity of the captions.

\subsection{Evaluation Results of EAH on General Vision-language Tasks and Benchmarks}

It is shown that in Table \ref{mme} and Table \ref{benchmark}, compared to the baseline model LLaVA1.5, our EAH method achieves non-negligible gains on all benchmark datasets without introducing additional computation during inferencing. Such performance improvements highlight the potential of EAH in enhancing LVLM's general visual perception capabilities.



                             

\subsection{Ablation Study}

\subsubsection{ Generalization Study of EAH on Other LVLMs}
\begin{table}[t]
\centering
\footnotesize
\scalebox{0.9}{
\begin{tabular}{@{}cccccc}
\toprule
 \multirow{2}{*}{Decoding} & \multicolumn{2}{c}{\textbf{Object-level}}                                   & \multicolumn{2}{c}{\textbf{Attribute-level}}                               & \multicolumn{1}{c}{\multirow{2}{*}{Total Scores$\uparrow$}} \\
 & \multicolumn{1}{c}{\textit{Existence}$\uparrow$} & \multicolumn{1}{c}{\textit{Count}$\uparrow$} & \multicolumn{1}{c}{\textit{Position}$\uparrow$} & \multicolumn{1}{c}{\textit{Color}$\uparrow$} & \multicolumn{1}{c}{}                       \\ \midrule
 Beam                    & 175.67 & 124.67 & 114.00 & 151.00 & 565.34 \\
 Greedy                    & 185.00 & 93.33 & 110.00 & 156.67 & 545.00 \\
 DOLA  \cite{dola}                  & 175.00 & 108.33 & 90.00 & 138.33 & 511.66 \\
 VCD   \cite{vcd}                    & \underline{184.66} & $\textbf{137.33}$ & \underline{128.67} & \underline{153.00} & \underline{603.66} \\
 OPERA   \cite{opera}                  & 180.67 & \underline{133.33} & 111.67 & 123.33 & 549.00 \\
 \textbf{Our}             & $\textbf{190.00}$ & 108.33 & $\textbf{145.00}$ & $\textbf{160.66}$ & $\textbf{603.99}$ \\ 
\bottomrule
\end{tabular}
}
\caption{Evaluation results on the hallucination subset of MME \cite{mme}. 
The best performances within each setting are \textbf{bolded}, max-tokens=512, baseline: LLaVA1.5-7B, $Layer=2$, $\beta=0.002$.}
\label{mme}
\end{table}

\begin{table}[t]
\centering
\footnotesize
\resizebox{8.5cm}{0.8cm}{
\setlength{\tabcolsep}{4pt} 
\begin{tabular}{cccccccc}
\toprule
Method & MM-Vet $\uparrow$ & VizWiz$\uparrow$ & Vqav2$\uparrow$ & Seed$\uparrow$ & GQA$\uparrow$& LLaVA-Bench$\uparrow$ & BLINK$\uparrow$  \\ \midrule
Baseline & 31.1 & 50.10 & 78.5 & 57.67 & 62.0 & 58.9 & 37.13 \\
VCD\cite{vcd} & 29.4 & 50.50 & - & 58.33 & 61.6 & 61.9 & - \\
Ours & \textbf{31.7} & \textbf{53.85} & \textbf{78.6} & \textbf{60.16} & \textbf{62.3} & \textbf{61.9} & \textbf{38.56} \\ \bottomrule
\end{tabular}
}
\caption{Evaluation results on general vision-language benchmarks, baseline: LLaVA1.5-7B, $Layer=2$, $\beta=0.002$.}
\label{benchmark}
\vspace{-0.5cm}
\end{table}


In contemporary MLLMs, images are processed by a CLIP model, mapped through different projectors, and integrated with large language models (LLMs). We hypothesize that the convergence of information flow in the early layers is affected by how different projectors—such as Linear, MLP, Cross-attention, and Q-former—map images to tokens. As shown in Table \ref{table3}, to test this hypothesis, we apply the EAH method to various models, including LLaVA1.5 \cite{LLaVA}, Shirka \cite{shikra}, Minigpt4 \cite{minigpt4}, Instructblip \cite{instructblip}, Minigemini \cite{minigemini}, QwenVL \cite{bai2023qwen} and InternVL \cite{internvl}. Notably, Shikra, LLaVA, Intern-VL, Qwen-VL, and Mini-Gemini use greedy search for decoding, while InstructBLIP uses beam search with a beam size of 5. Despite the different decoding strategies and projectors, all models exhibit a consistent pattern of dense attention sink in the shallow layers and sparse attention sink in the deeper layers. Applying EAH to these models consistently improves performance, demonstrating its effective plug-and-play capability and broad applicability.

Moreover, we find that models using MLP or Linear projectors, such as LLaVA, Intern-VL, and Shikra, show the most significant improvements after applying EAH. In contrast, models with cross-attention or Q-former projectors, like Qwen-VL and MiniGPT-4, exhibit modest gains. 

In the \textbf{supplementary material}, we provide several attention maps for different LVLMs. It is observed that some models, such as MiniGPT-4 and Qwen-VL, also display a pattern of dense attention sinks in shallow layers and sparse attention sinks in deep layers. However, the gap between shallow and deep layer attention sinks is less pronounced compared to models like LLaVA1.5 and Intern-VL, resulting in relatively smaller improvements. Given that most current LVLMs are fine-tuned versions of LLaMA (vicuna) \cite{llama} or Qwen \cite{qwen2}, the attention map distribution in LVLMs likely inherits patterns from the underlying LLMs. Therefore, we also visualize Qwen and LLaMA with their attention maps. We find a consistent pattern in these LLMs: dense attention sinks in shallow layers and sparse attention sinks in deeper layers. When examining individual attention heads, we observe that certain attention head distributions in LVLMs and LLMs are quite similar.

\begin{table}[t]
  \centering
  \scalebox{0.8}{
  \begin{tabular}{cc|ccc}
    \toprule
    Model & Projector & CHAIR\( _{S} \) $\downarrow$ & CHAIR\( _{I} \) $\downarrow$ \\
    \midrule
    MiniGPT-4 \cite{minigpt4} & Qformer & 31.8 & 9.9 \\
    Instructblip \cite{instructblip} & Qformer & 58.8 & 23.7 \\
    Shikra \cite{shikra} & Linear & 55.8 & 15.4 \\
    QwenVL-Chat \cite{bai2023qwen} & Cross-attention & 45.6 & 12.5 \\
    LLaVA1.5 \cite{LLaVA} & MLP & 47.0 & 13.8 \\
    Mini-Gemini \cite{minigemini} & MLP & 32.6 & 8.7 \\
    InternVL \cite{internvl} & MLP & 45.8 & 12.9 \\
    \midrule
   MiniGPT-4+\textbf{EAH} & Qformer & 30.4(\textcolor{blue}{-1.4}) & 9.5(\textcolor{blue}{-0.4}) \\ 
    Instructblip+\textbf{EAH} & Qformer & 56.0((\textcolor{blue}{-2.8}) & 15.7((\textcolor{blue}{-8.0}) \\
    Shikra+\textbf{EAH} & Linear  & 47.9((\textcolor{blue}{-7.9}) & 13.7((\textcolor{blue}{-1.7}) \\
    QwenVL-Chat+\textbf{EAH} & Cross-attention & 44.6((\textcolor{blue}{-1.0}) & 11.9((\textcolor{blue}{-0.6}) \\
    LLaVA1.5+\textbf{EAH} & MLP & 36.4((\textcolor{blue}{-10.6}) & 9.9((\textcolor{blue}{-3.8}) \\
    Mini-Gemini+\textbf{EAH} & MLP & 27.8((\textcolor{blue}{-4.8}) & 8.5((\textcolor{blue}{-0.2}) \\
    InternVL+\textbf{EAH} & MLP & 32.4((\textcolor{blue}{-13.4}) & 9.0((\textcolor{blue}{-3.9}) \\
    \bottomrule
  \end{tabular}}
    \caption{Generalization study of EAH on other LVLMs models.}
      \label{table3}
      \vspace{-0.2cm}
\end{table}

\subsubsection{ Generalization Study of EAH on LLM Models}

As mentioned above, we find a similar pattern in LLMs and LVLMs. To verify the feasibility of EAH on LLMs, we chose four models including LLaMA3.1-instruct \cite{llama3.1}, Ministral-8B-Inst \cite{mistral}, Qwen-2-7B-Inst \cite{qwen2} and Qwen-2.5-7B-Inst \cite{qwen2.5}. The datasets are GSM8K \cite{gsm8k} and TruthfulQA \cite{truthfulqa}, respectively. The results are shown in Table.\ref{llm-table}, which demonstrates that EAH can produce consistency gains in LLM as well. 

\begin{table}[t]
\centering
\label{tab:prompts-llama3}
\scalebox{0.75}{
\begin{tabular}{@{}llccc@{}}
\toprule
Model & Dataset & Metric & Baseline & \textbf{w/ EAH} \\ 
\midrule
\multirow{2}{*}{Llama-3.1-8B-Inst\cite{llama3.1}  } 
& \multirow{1}{*}{GSM8K} & Acc $\uparrow$ & 85.29  & \textbf{87.25}((\textcolor{blue}{+1.96}) \\ 
& \multirow{1}{*}{Truthful-QA} & Acc $\uparrow$ & 49.27  & \textbf{53.17}((\textcolor{blue}{+3.90}) \\ 

\midrule
\multirow{2}{*}{Ministral-8B-Inst \cite{mistral} } 
& \multirow{1}{*}{GSM8K} & Acc $\uparrow$ & 90.00  & \textbf{91.36}((\textcolor{blue}{+1.36}) \\ 
& \multirow{1}{*}{Truthful-QA} & Acc $\uparrow$ &  47.80  & \textbf{51.22}((\textcolor{blue}{+3.42}) \\ 

\midrule
\multirow{2}{*}{Qwen-2-7B-Inst \cite{qwen2} } 
& \multirow{1}{*}{GSM8K} & Acc $\uparrow$ & 88.63 & \textbf{89.22}((\textcolor{blue}{+0.59}) \\ 
& \multirow{1}{*}{Truthful-QA} & Acc $\uparrow$ & 45.85 & \textbf{46.34}((\textcolor{blue}{+0.49}) \\

\midrule
\multirow{2}{*}{Qwen-2.5-7B-Inst \cite{qwen2.5} } 
& \multirow{1}{*}{GSM8K} & Acc $\uparrow$ & 92.72 & \textbf{93.63}((\textcolor{blue}{+0.91}) \\ 
& \multirow{1}{*}{Truthful-QA} & Acc $\uparrow$ & 52.68  & \textbf{56.10}((\textcolor{blue}{+3.42}) \\ 

\bottomrule
\end{tabular}
}
\caption{Generation study of EAH on LLM models. }
\label{llm-table}
\end{table}
\subsubsection{Ablation Study of Hyper-parameter}


Table \ref{table:layer_head_results} presents the ablation study results for the parameters \textbf{Threshold}: $\beta$, \textbf{Layer}: $L$, \textbf{Top Head}: $\mathcal{N}$. The experimental results indicate that the configuration with layer=2, $\beta$=0.002, and $\mathcal{N}$=top1, yields the best performance, achieving \textbf{C\( _{S} \)} of 36.6 and \textbf{C\( _{I} \)} of 9.9.

The improvement achieved by broadcasting the top attention head primarily benefits from the centralization of attention. Since most vision sinks are concentrated in the first and second layers, broadcasting the attention map of the head with the densest vision sink in these layers to the other heads helps unify each head's focus on visual information, forming a "consensus" attention pattern. Ultimately, the high-density vision sink pattern enables the model to capture key information from the image, effectively reducing hallucinations.

\begin{table}[t]
\hspace{-11.5pt}
\centering
\renewcommand{\arraystretch}{0.9} 
\setlength{\tabcolsep}{2pt} 
\scalebox{0.858}{ 
\begin{tabular}{@{}ccc@{}} 
    \begin{subtable}[t]{0.18\textwidth}
        \centering
        \begin{tabular}{cc|cc}
        \toprule
        \textbf{$\mathcal{N}$} & \textbf{$\beta$} & \textbf{C\( _{S} \)}$\downarrow$ & \textbf{C\( _{I} \)}$\downarrow$ \\ 
        \midrule
        \multirow{2}{*}{1} & 0.0015 & \textbf{39.4} & \textbf{9.9} \\ 
                           & 0.0020  & 41.8 & 10.4 \\ 
        \cmidrule(lr){1-4}
        \multirow{2}{*}{2} & 0.0015 &  \underline{41.2} &  \underline{10.3} \\ 
                           & 0.0020  & 42.2 & 10.9\\ 
        \cmidrule(lr){1-4}
        \multirow{2}{*}{3} & 0.0015 & 41.6 &  11.2\\ 
                           & 0.0020  & 42.0 & 11.0 \\ 
        \cmidrule(lr){1-4}
        R & -- & 44.2 & 12.0 \\ 
        \bottomrule
        \end{tabular}
        \caption{$Layer=1$}
    \end{subtable}
    &
    \begin{subtable}[t]{0.18\textwidth}
        \centering
        \begin{tabular}{cc|cc}
        \toprule
        \textbf{$\mathcal{N}$} & \textbf{$\beta$} & \textbf{C\( _{S} \)}$\downarrow$ & \textbf{C\( _{I} \)}$\downarrow$ \\ 
        \midrule
        \multirow{2}{*}{1} & 0.0015 & 40.6 & 11.9 \\ 
                           & 0.0020  & \textbf{36.4} & \textbf{9.9} \\ 
        \cmidrule(lr){1-4}
        \multirow{2}{*}{2} & 0.0015 &\underline{40.6}&  \underline{11.7} \\ 
                           & 0.0020  &  42.4 & 11.9 \\ 
        \cmidrule(lr){1-4}
        \multirow{2}{*}{3} & 0.0015 & 44.4 & 12.0 \\ 
                           & 0.0020  & 44.2 & 12.5 \\ 
        \cmidrule(lr){1-4}
        R & -- & 42.8 & 12.4 \\ 
        \bottomrule
        \end{tabular}
        \caption{$Layer=2$}
    \end{subtable}
    &
    \begin{subtable}[t]{0.18\textwidth}
        \centering
        \begin{tabular}{cc|cc}
        \toprule
        \textbf{$\mathcal{N}$} & \textbf{$\beta$} & \textbf{C\( _{S} \)}$\downarrow$ & \textbf{C\( _{I} \)}$\downarrow$ \\ 
        \midrule
        \multirow{2}{*}{1} & 0.0015 & 49.0 & 14.3 \\ 
                           & 0.0020  & \textbf{46.8} &  \underline{13.8} \\ 
        \cmidrule(lr){1-4}
        \multirow{2}{*}{2} & 0.0015 & 49.8 & 14.1 \\ 
                           & 0.0020  & 49.6 & 14.0 \\ 
        \cmidrule(lr){1-4}
        \multirow{2}{*}{3} & 0.0015 & 48.4 & 13.8 \\ 
                           & 0.0020  &  \underline{48.2} &  \textbf{13.7} \\ 
        \cmidrule(lr){1-4}
        R & -- & 49.8 & 14.5 \\ 
        \bottomrule
        \end{tabular}
        \caption{$Layer=3$}
    \end{subtable}
\end{tabular}
}
\caption{Results for ablation study of the hyperparameter on CHAIR dataset \cite{chair}, Threshold: $\beta$, $\mathcal{N}$: broadcast top $\mathcal{N}$ head, baseline: LLaVA1.5-7B.}
\label{table:layer_head_results}
\end{table}

\subsubsection{Grad-CAM Results of EAH }

\begin{figure}[t]
\vspace{-0.4cm}
\centerline{\includegraphics[scale=0.33]{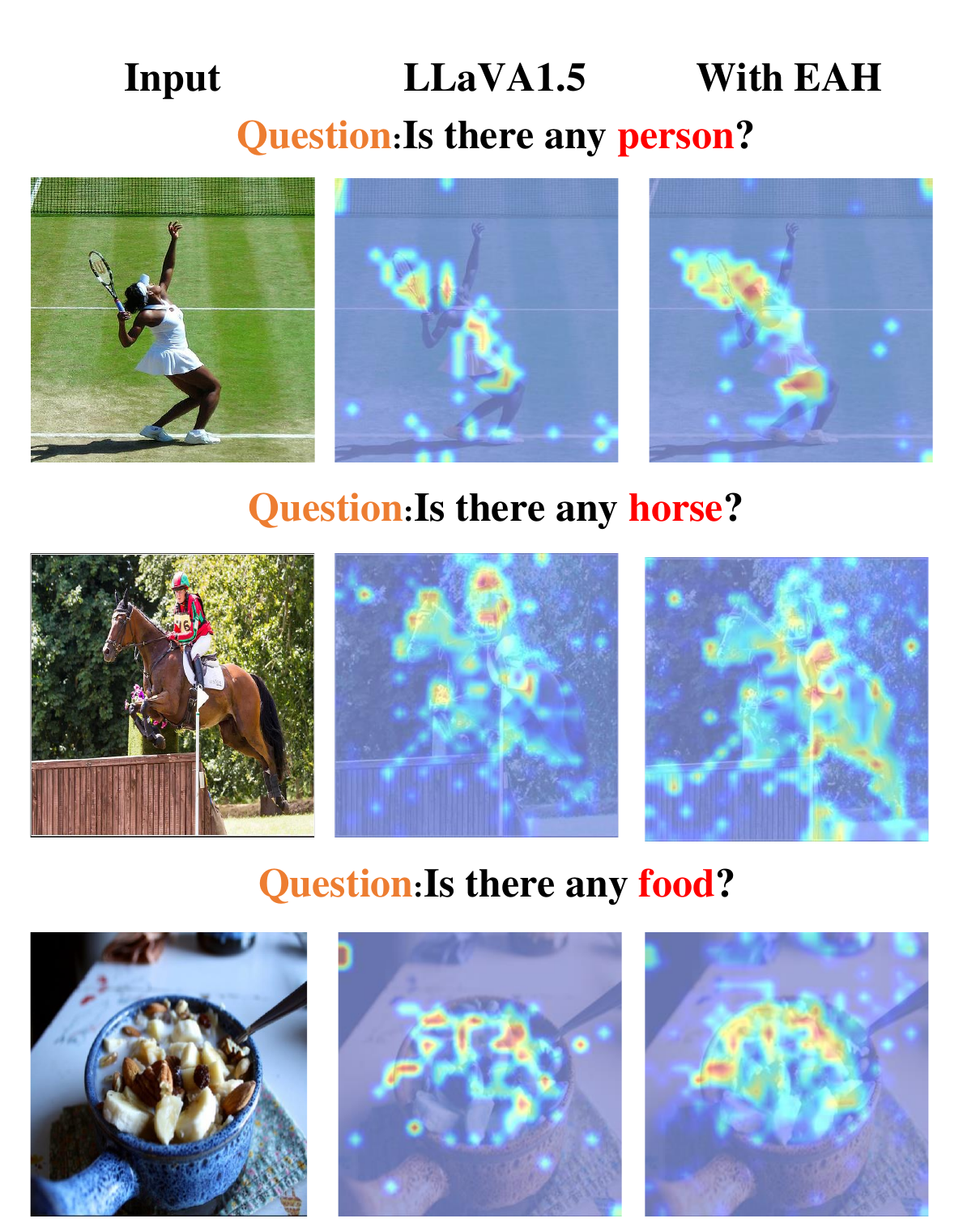}}
\caption{The Grad-CAM results of LLaVA1.5 and LLaVA1.5 after adding EAH.  }
\label{grad-cam}

\end{figure}

At the same time, we demonstrate the heat map of the model through LLaVA-CAM \cite{LLaVA-CAM}. As shown in Fig. \ref{grad-cam}, the comparison between the base model LLaVA1.5 without EAH and with EAH using Grad-CAM visualizations reveals that the model with EAH shows slightly increased attention to key objects in the image at shallow layers, such as "horse," "person," and "food". 
This result demonstrates that the EAH method, which enhances attention heads in shallow layers, improves the model's generalization ability. It enables the model to focus more on essential regions in the image, strengthening the flow of information and enhancing its overall capabilities.

\section{ Conclusion}
In this paper, we introduce a training-free and plug-and-play method named \textbf{E}nhancing \textbf{A}ttention \textbf{H}eads (EAH) to alleviate the challenge of hallucinations in multimodal language models (MLLMs). EAH is designed to enhance the densities and distribution of image token attention sinks in the shallow layers, thereby mitigating hallucinations. Our extensive benchmark tests on hallucination and generalization experiments demonstrate the plug-and-play effectiveness of EAH as a training-free approach.

{
    \small
    \bibliographystyle{ieeenat_fullname}
    \bibliography{main}
}
\newpage
\appendix
\section{Appendix}
\subsection{Discussion and Limitations}

The results of this paper validate ITI's \cite{ITI} conclusion that only a subset of attention heads plays a significantly more prominent role. Effectively optimizing these key attention heads is likely to yield substantial improvements in model efficiency and overall performance. While the method proposed in this paper significantly alleviates hallucinations and demonstrates both simplicity and effectiveness, it still has some limitations. For instance, it exhibits about a 4 percentage point lower recall compared to other methods at the same $avg_{len}$. However, it is also important to note that EAH outperforms greedy search methods to maintain diversity and serves as a robust plug-and-play and training-free solution. To address hallucination issues more fundamentally, we believe that improved alignment of projectors and advanced training methods, such as reinforcement learning from human feedback (RLHF), is necessary for more effective resolution.

\subsection{Why EAH in Q/K before V}
The reason for intervening before V is that the attention matrix attn\_weights (i.e., attention\_map) represents the weight distribution between different queries and keys. EAH (Enhancing Attention Heads) specifically modifies these weights to adjust the attention distribution. If the intervention happens before V, it allows direct control over the attention concentration during the soft weight allocation stage, making attn\_weights more focused on the relevant image information. The adjusted attn\_weights, when multiplied with V, will more effectively filter out important information.

If the intervention occurs after V, the effect of EAH will be limited to the final attn\_output value, rather than modifying the attention matrix itself. This makes it harder to effectively control the attention on specific tokens.

Therefore, intervening at the attn\_weights stage before V allows for a more direct impact on the model's focus on different tokens, thereby improving performance.
\begin{figure}[t]
\centerline{\includegraphics[scale=0.5]{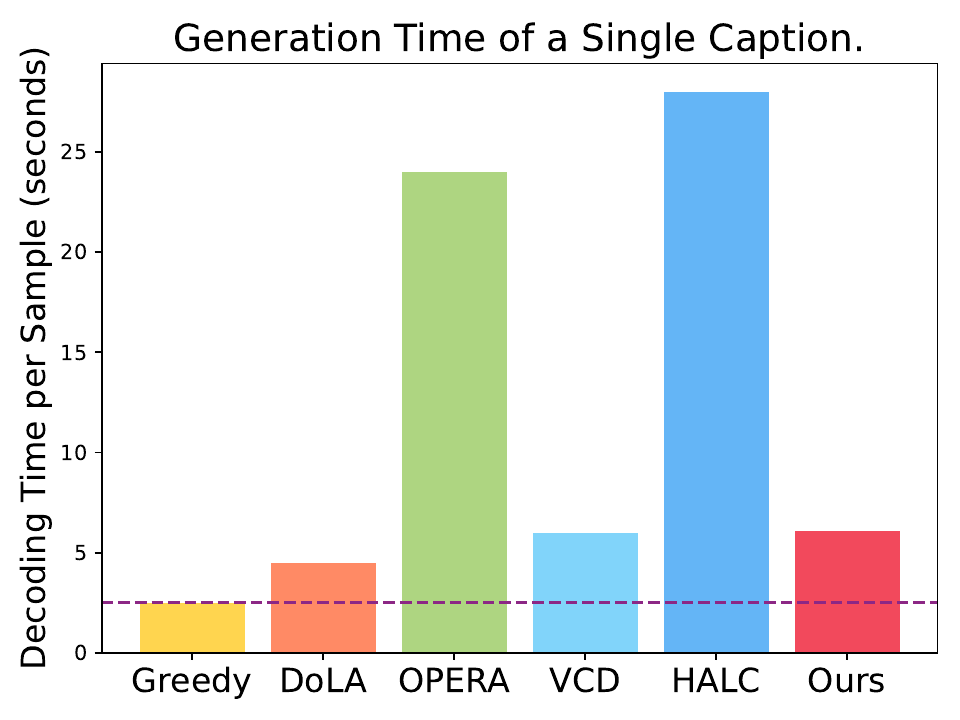}
}
\caption{Generation time of a single response.}
\label{time}
\end{figure}


\begin{figure*}[h]
\centerline{\includegraphics[scale=0.22]{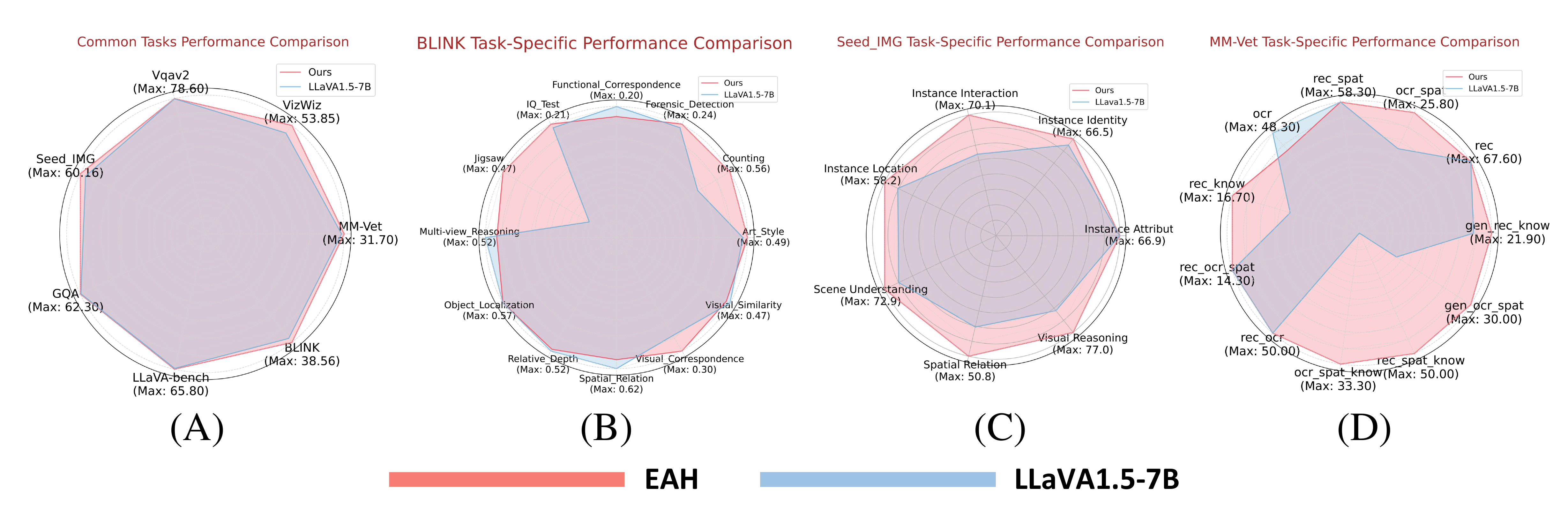}}
\caption{The generalization results on the datasets for the seven multimodal models (MMEs) demonstrate that EAH combined with LLaVA1.5 enhances the metrics compared to LLaVA1.5 alone. This not only highlights the exceptional performance of EAH in addressing hallucinations but also reinforces the notion that EAH can generally enhance the model's overall performance. }
\label{leida}
\end{figure*}

\subsection{More ablation study}
\subsubsection{Comparison of Generation Time}

In Fig. \ref{time}, we compare the generation time of EAH with existing methods for alleviating hallucinations. Both EAH and OPERA \cite{opera} are methods that require attention intervention, and we utilize a standard self-attention implementation. In contrast, other methods such as Greedy, DoLA, VCD, and HALC \cite{halc} do not necessitate attention intervention. All methods were tested on a single A100-80GB GPU. Our observations indicate that EAH achieves a decoding time similar to that of VCD \cite{vcd}. It is slightly longer than the Greedy and DoLA \cite{dola} methods due to our intervention in the attention weights at layer 1.2 during inference. In comparison, the other methods inevitably introduce additional computational overhead.

\subsubsection{Qualitative Experiment of Thresholds and Layers}

Table \ref{table2} presents the results of our ablation experiments, which assess the impact of various thresholds and layers on model performance. We test different thresholds (0.0006, 0.0008, 0.0015, 0.002) and layers (1, 2, 3, 4, 16, 32) to observe their effects on the model's performance with the CHAIR dataset.

The results show that the model achieves its highest performance on the CHAIR$_s$ and CHAIR$_I$ metrics, with scores of 36.6 and 10.0, respectively, when applying EAH at the second layer with a threshold of 0.002. However, both metrics significantly decrease as the number of layers increases. This supports our hypothesis that information flow converges in early layers and diverges in deeper layers, and keeping the attention sink dense in shallow layers will effectively alleviate the hallucination. As the depth increases, both CHAIR$_s$ and CHAIR$_I$ values rise and exceed the baseline, suggesting that the likelihood of the model generating hallucinations at deeper layers increases. This occurs because attention sinks become more sparse in deeper layers and the differences between attention heads diminish. Therefore, even if the most densely concentrated attention-sink head is identified and broadcasted to other heads, its impact may still be limited.

\subsubsection{Experiment results on other hallucination benchmark}
\begin{table}[h!]
\centering
\resizebox{8cm}{0.93cm}{
\begin{tabular}{ccccc}
\toprule
\multicolumn{5}{c}{\textbf{HallusionBench \cite{hallusionbench}}} \\
\midrule
\textbf{split} & \textbf{method} & \textbf{aAcc} $\uparrow$ & \textbf{fAcc} $\uparrow$& \textbf{qAcc}$\uparrow$ \\
\midrule
\multirow{2}{*}{Overall} & LLaVA1.5-7B & 35.54154 & 17.63006 & 11.20879 \\
                         & EAH         & 44.37434 & 17.91908 & 14.50549 \\
\bottomrule
\end{tabular}}
\end{table}

\begin{table}[t]
  \caption{Qualitative experiment of layer and threshold on CHAIR dataset (baseline: LLaVA-1.5-7B, head=top1).}
  \centering
  \resizebox{8cm}{4cm}{
  \begin{tabular}{cc|cccccc}
    \toprule
    Layer & Threshold & CHAIR\( _{S} \) $\downarrow$ & CHAIR\( _{I} \) $\downarrow$ & Recall $\uparrow$ & Avg. Len \\
    \midrule
    1 & 0.0006  & 46.4  & 13.9 & 76.9 & 96.1 \\
    1 & 0.0008  & 46.0  & 12.9 & 77.1 & 101.2 \\
    1 & 0.0015  & \underline{39.4}  & \textbf{9.9} & 72.2 & 108.6 \\
    1 & 0.002   & 41.8  & 10.4 & 72.9 & 112.4 \\ 
    2 & 0.0006  & 41.4  & 12.3 & 76.4 & 95.6 \\
    2 & 0.0008  & 43.0  & 11.6 & 74.9 & 100.9 \\
    2 & 0.0015  & 40.6  & 11.9 & 74.9 & 102.7  \\
    2 & 0.002   & \textbf{36.4}  & \underline{9.9} & 73.9 & 97.7  \\  
    3 & 0.0006  & 49.0  & 14.3 & 78.0 & 98.6 \\
    3 & 0.0008  & 49.4  & 14.3 & 78.3 & 98.4 \\
    3 & 0.0015  & 49.0  & 14.3 & 77.9 & 98.5 \\
    3 & 0.002   & 46.8  & 13.8 & \textbf{78.5} & 98.7 \\  
    4 & 0.0006  & 46.4  & 13.6 & 76.9 & 96.1 \\
    4 & 0.0008  & 50.0  & 14.7 & 78.3 & 97.6 \\
    4 & 0.0012  & 49.6  & 14.6 & \underline{78.4} & 97.7 \\
    4 & 0.002   & 49.4  & 14.5 & 78.2 & 97.5 \\  
    16 & 0.0006 & 53.0  & 14.8 & 77.9 & 100.9 \\
    16 & 0.0008 & 49.2  & 14.8 & 77.4 & 100.5 \\
    16 & 0.0015 & 52.6  & 15.0 & 78.0 & 100.9 \\
    16 & 0.002  & 47.2  & 14.1 & 77.3 & 98.7 \\  
    32 & 0.0006 & 50.8  & 14.4 & 78.1 & 98.7 \\
    32 & 0.008  & 50.8  & 14.4 & 78.1 & 98.7 \\
    32 & 0.0015 & 47.0  & 13.8 & 76.9 & 95.8 \\
    32 & 0.002  & 53.0  & 14.8 & 77.9 & 100.9 \\
    \bottomrule
  \end{tabular}}
  \label{table2}
\end{table}

\subsubsection{Qualitative Experiment of Heads Number}
As demonstrated in the previous section, the first two layers contain the most attention sinks. Therefore, we focus on applying the EAH strategy to these layers. In Table \ref{table3}, we present a qualitative experiment to assess the impact of increasing the number of attention heads affected. We test this by broadcasting the densest attention head across 4, 8, 16, and 32 heads. For instance, when broadcasting to 4 heads, the attention map from the densest head is duplicated across these 4 heads, while the remaining 28 heads remain unchanged.

The results indicate that broadcasting the densest attention head to 32 heads achieves the best performance. This suggests that using the densest attention pattern from the early layers improves the model’s focus on image information, enabling the model to concentrate on global image information rather than allowing attention to converge on specific tokens. This approach significantly helps to alleviate hallucinations.


\begin{table}[t]
  \caption{Qualitative experiment of enhancing head number on CHAIR dataset (baseline: LLaVA-1.5-7B).}
  \centering
  \resizebox{8cm}{2cm}{
  \begin{tabular}{cc|cccccc}
    \toprule
    Layer & Copy Head & CHAIR\( _{S} \) $\downarrow$ & CHAIR\( _{I} \) $\downarrow$ & Recall $\uparrow$ & Avg. Len \\
    \midrule
    1 & 4  & 51.0 & 14.7 & 78.1 & 98.9 \\
    1 & 8  & 49.4 & 14.5 & \textbf{78.5} & 99.5 \\
    1 & 16 & 48.4 & 13.5 & 77.1 & 99.8 \\ 
    1 & 28 & 40.6 & 11.8 & 74.1 & 103.8 \\
    1 & 32 & \underline{39.4} & \textbf{9.9} & 72.2 & 108.6 \\
    \midrule
    2 & 4  & 50.8 & 14.4 & 78.1 & 98.7 \\
    2 & 8  & 49.6 & 13.7 & 77.6 & 100.0 \\
    2 & 16 & 47.4 & 14.0 & 77.5 & 100.5 \\ 
    2 & 28 & 42.2 & 11.9 & 74.9 & 98.1 \\
    2 & 32 & \textbf{36.4} & \underline{9.9} & 73.9 & 97.7 \\
    \bottomrule
  \end{tabular}}
  \label{table3}
\end{table}

\begin{figure*}[h]
\centerline{\includegraphics[scale=0.25]{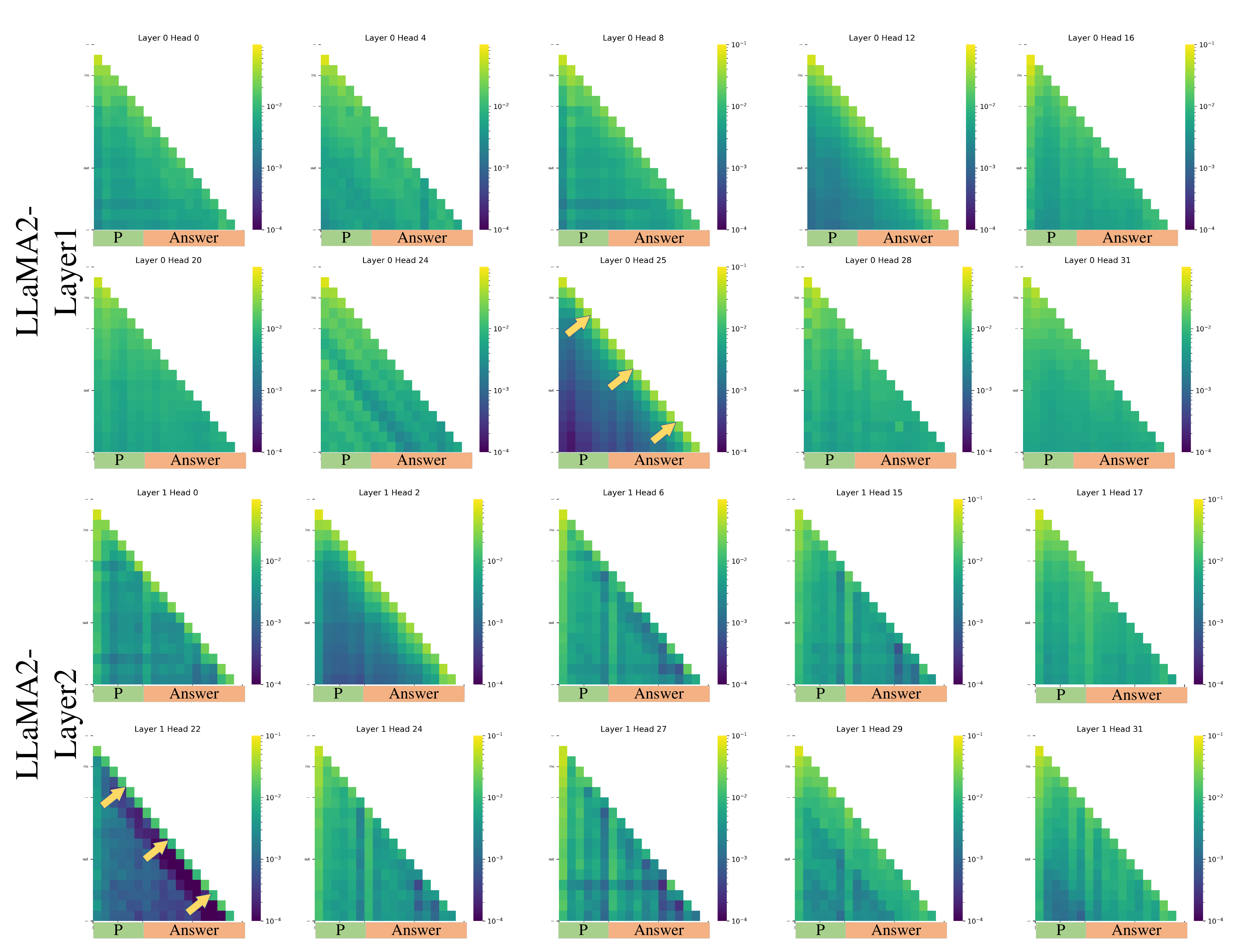}}
\caption{The flow map of EAH which aims to find the head with the densest attention sink in the 32 layers and broadcast the attention map of this head to other heads.}
\label{structure}
\end{figure*}

\begin{figure*}[t]
\centerline{\includegraphics[scale=0.25]{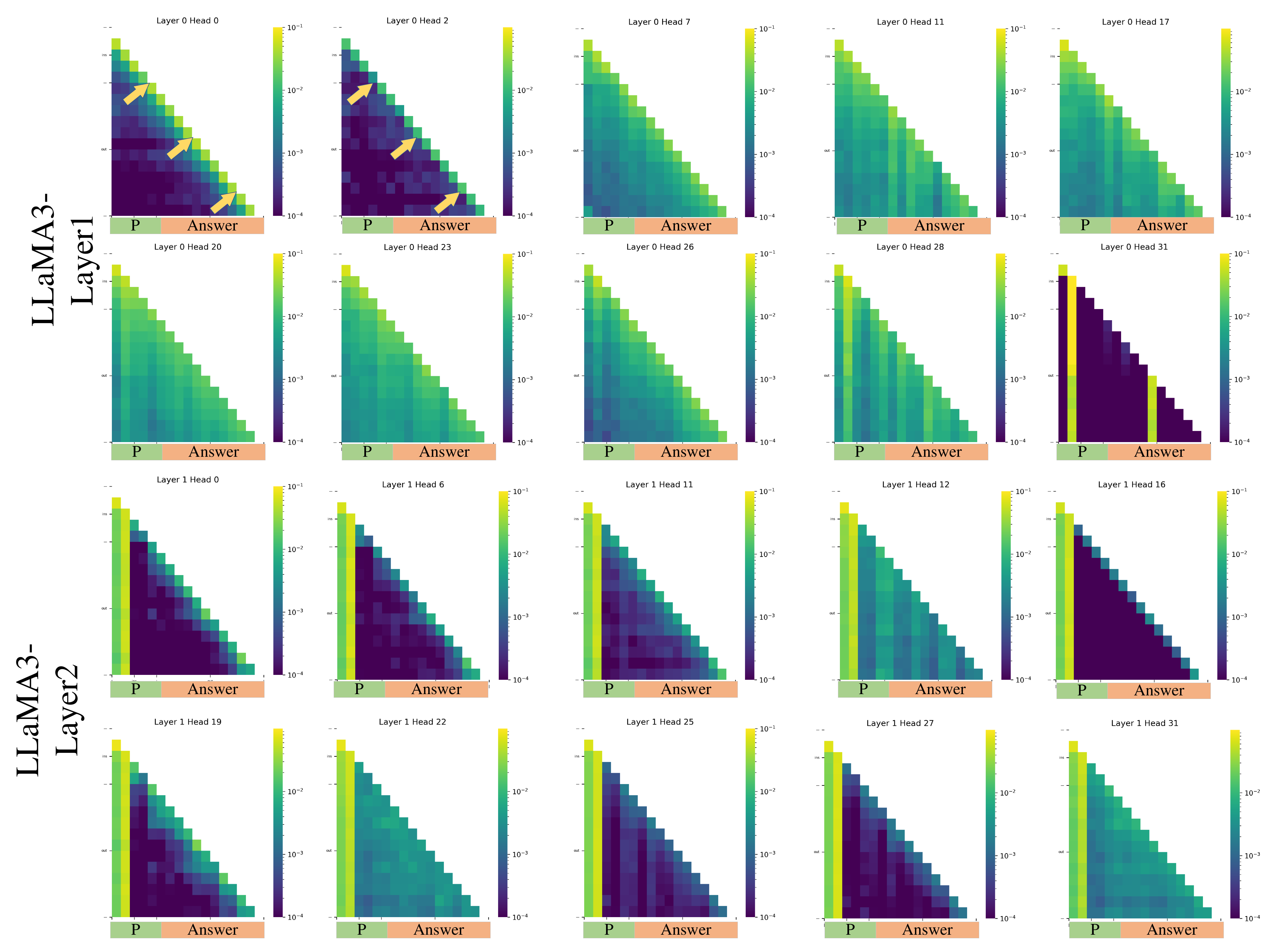}}
\caption{The flow map of EAH which aims to find the head with the densest attention sink in the 32 layers and broadcast the attention map of this head to other heads.}
\label{structure}
\end{figure*}

\begin{figure*}[t]
\centerline{\includegraphics[scale=0.25]{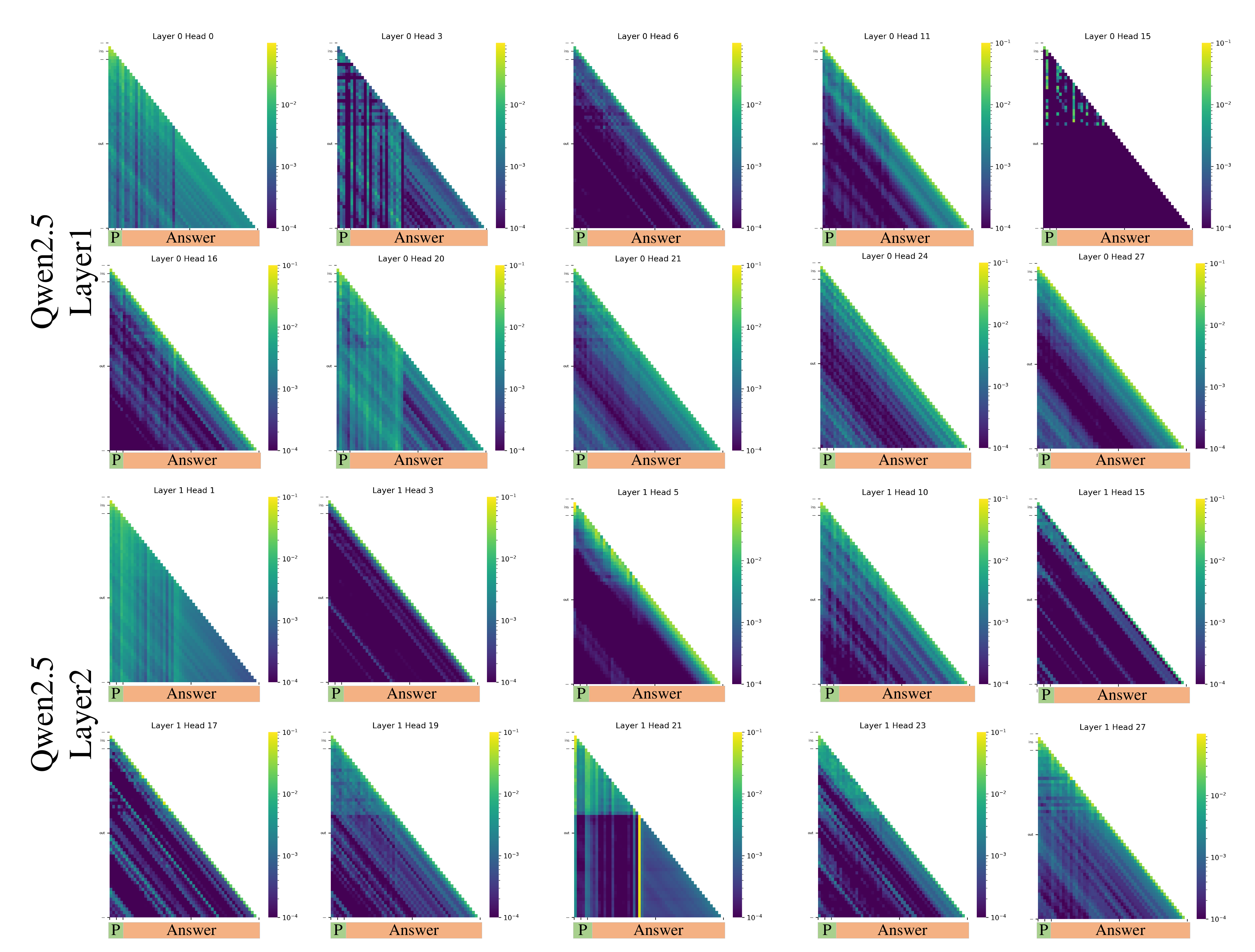}}
\caption{The flow map of EAH which aims to find the head with the densest attention sink in the 32 layers and broadcast the attention map of this head to other heads.}
\label{structure}
\end{figure*}

\begin{figure*}[t]
\centerline{\includegraphics[scale=0.25]{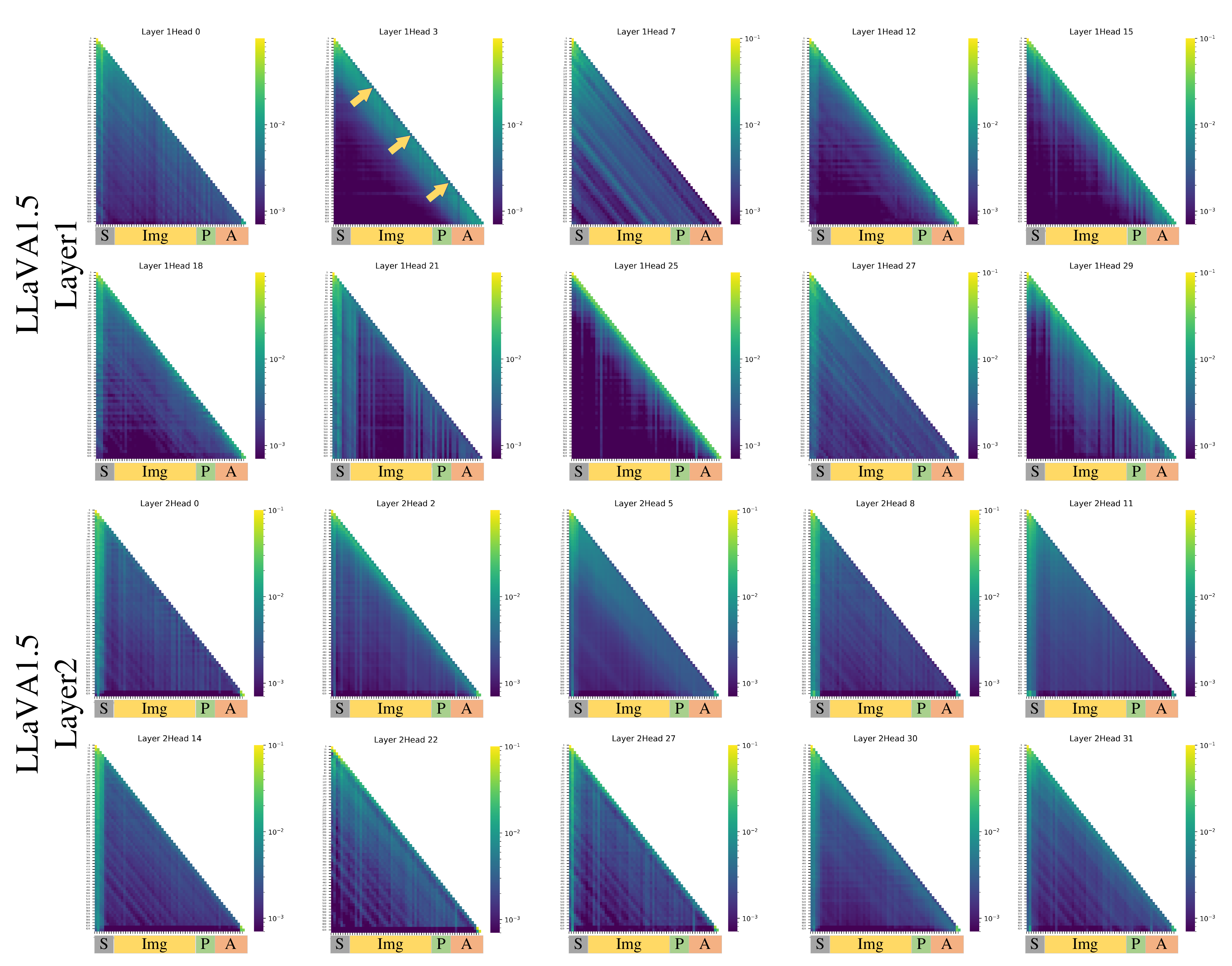}}
\caption{The flow map of EAH which aims to find the head with the densest attention sink in the 32 layers and broadcast the attention map of this head to other heads.}
\label{structure}
\end{figure*}

\begin{figure*}[t]
\centerline{\includegraphics[scale=0.25]{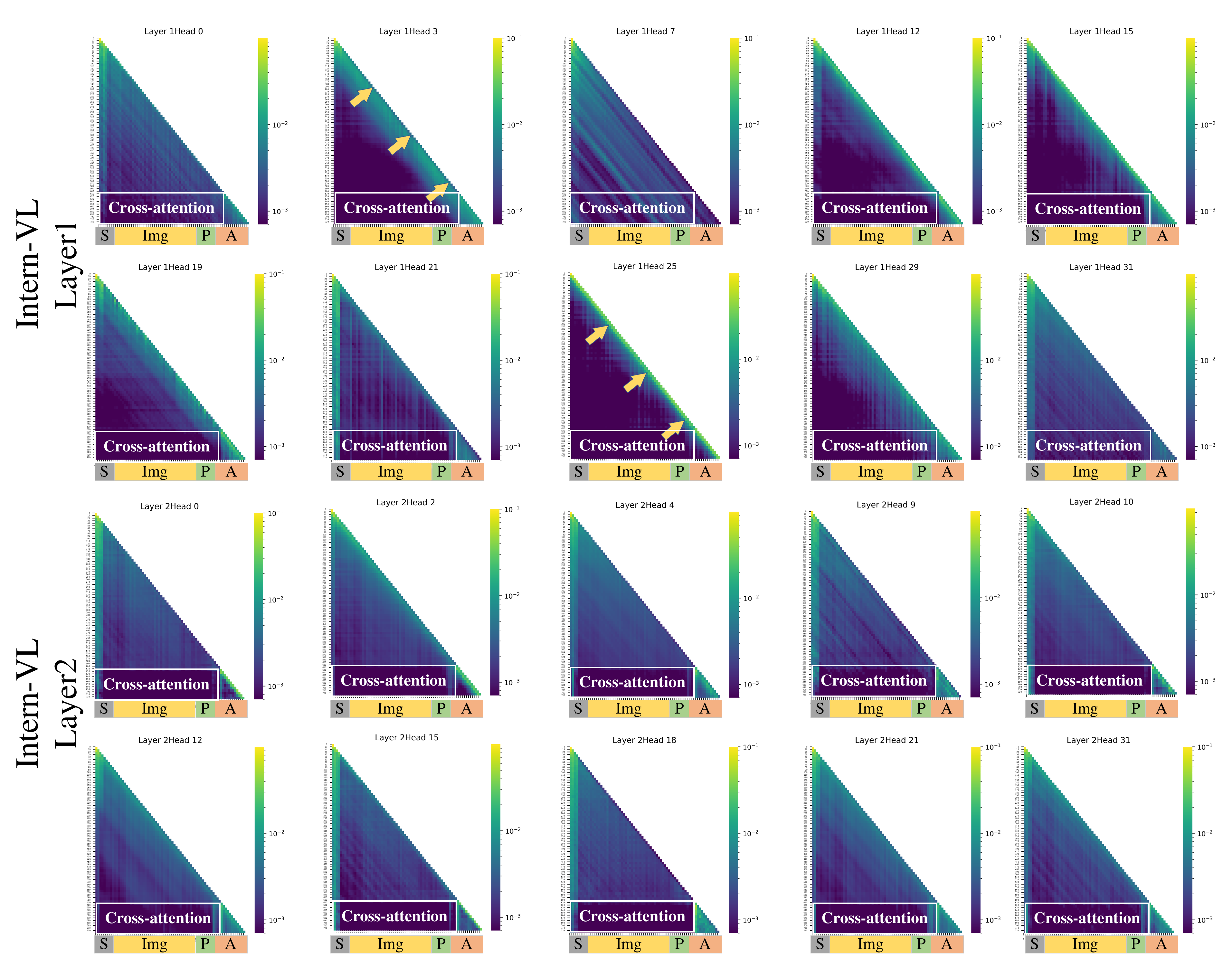}}
\caption{The flow map of EAH which aims to find the head with the densest attention sink in the 32 layers and broadcast the attention map of this head to other heads.}
\label{structure}
\end{figure*}

\begin{figure*}[t]
\centerline{\includegraphics[scale=0.25]{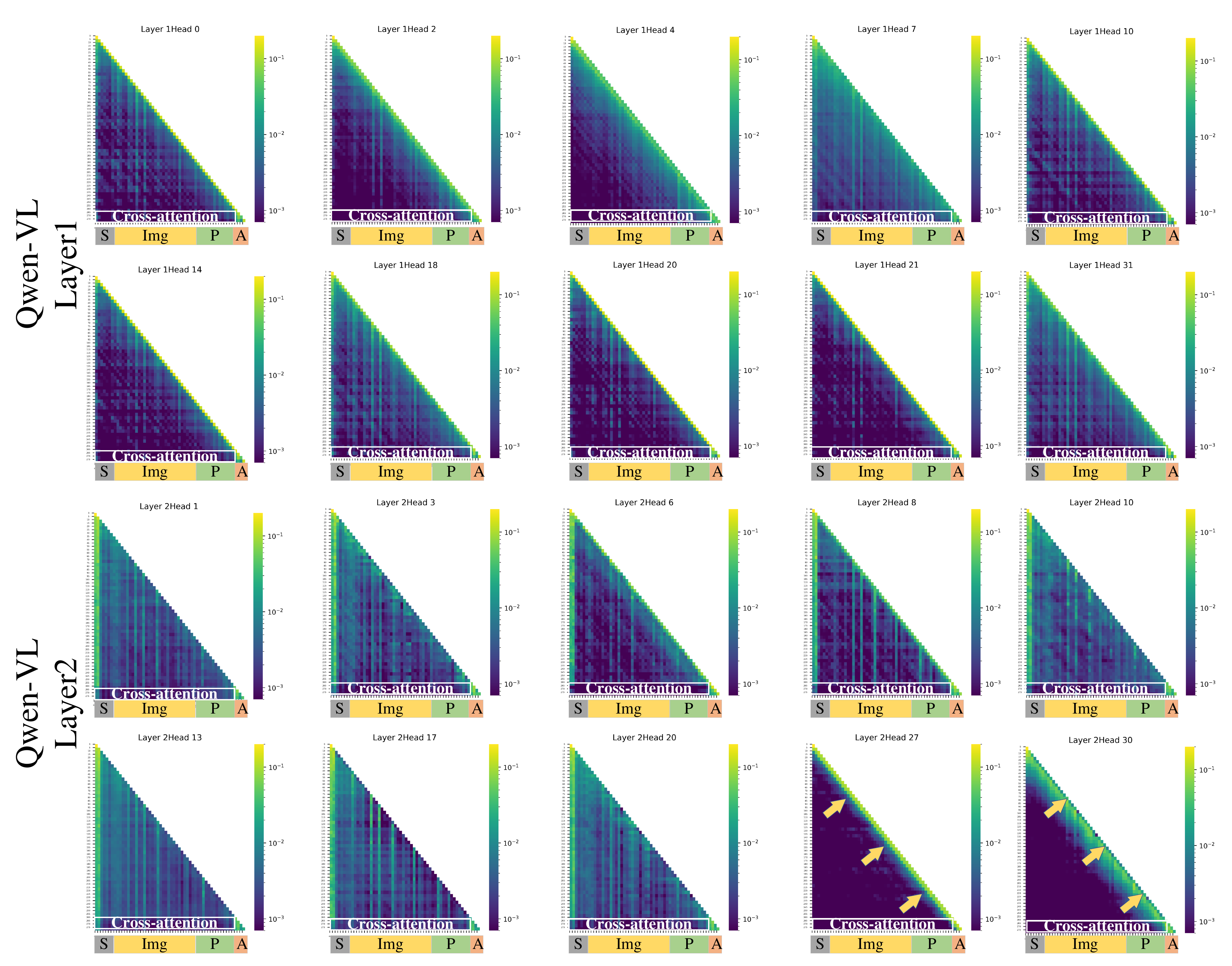}}
\caption{The flow map of EAH which aims to find the head with the densest attention sink in the 32 layers and broadcast the attention map of this head to other heads.}
\label{structure}
\end{figure*}

\begin{figure*}[t]
\centerline{\includegraphics[scale=0.25]{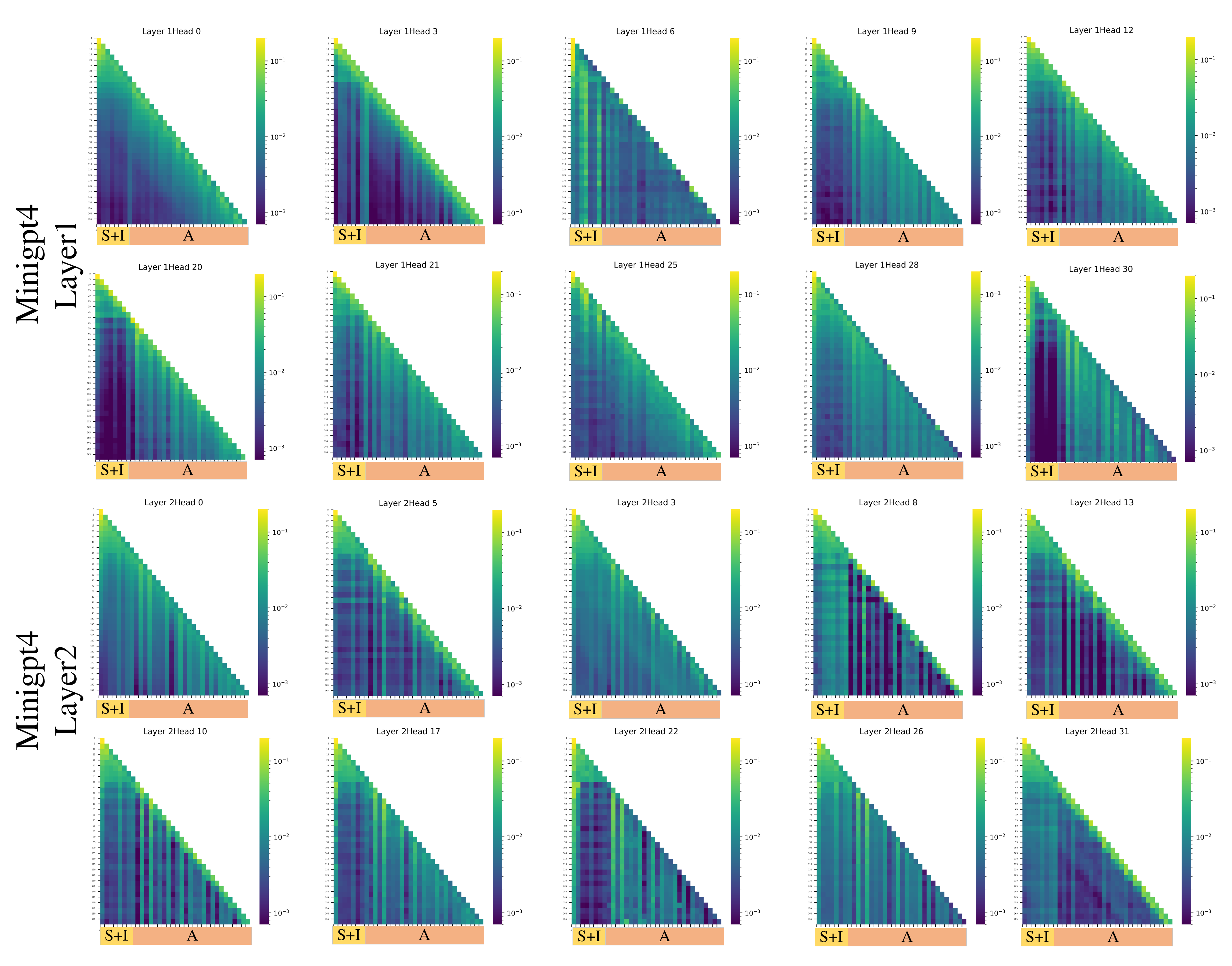}}
\caption{The flow map of EAH which aims to find the head with the densest attention sink in the 32 layers and broadcast the attention map of this head to other heads.}
\label{structure}
\end{figure*}

\begin{figure*}[h]
\centerline{\includegraphics[scale=0.37]{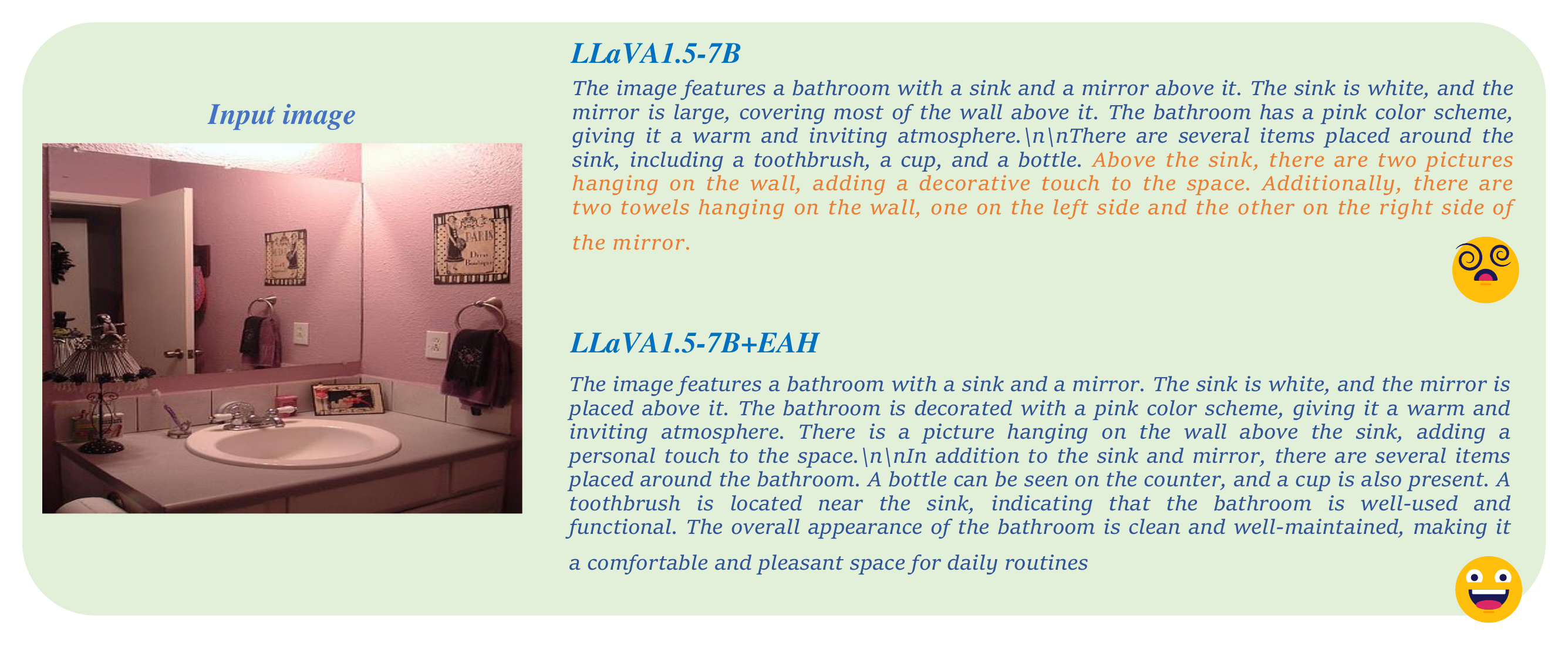}}
\caption{Results of LLaVA1.5 with EAH, EAH can  significantly reduce hallucinations while maintaining the original sentence length.}
\label{mgm+eah}
\end{figure*}

\begin{figure*}[h]
\centerline{\includegraphics[scale=0.37]{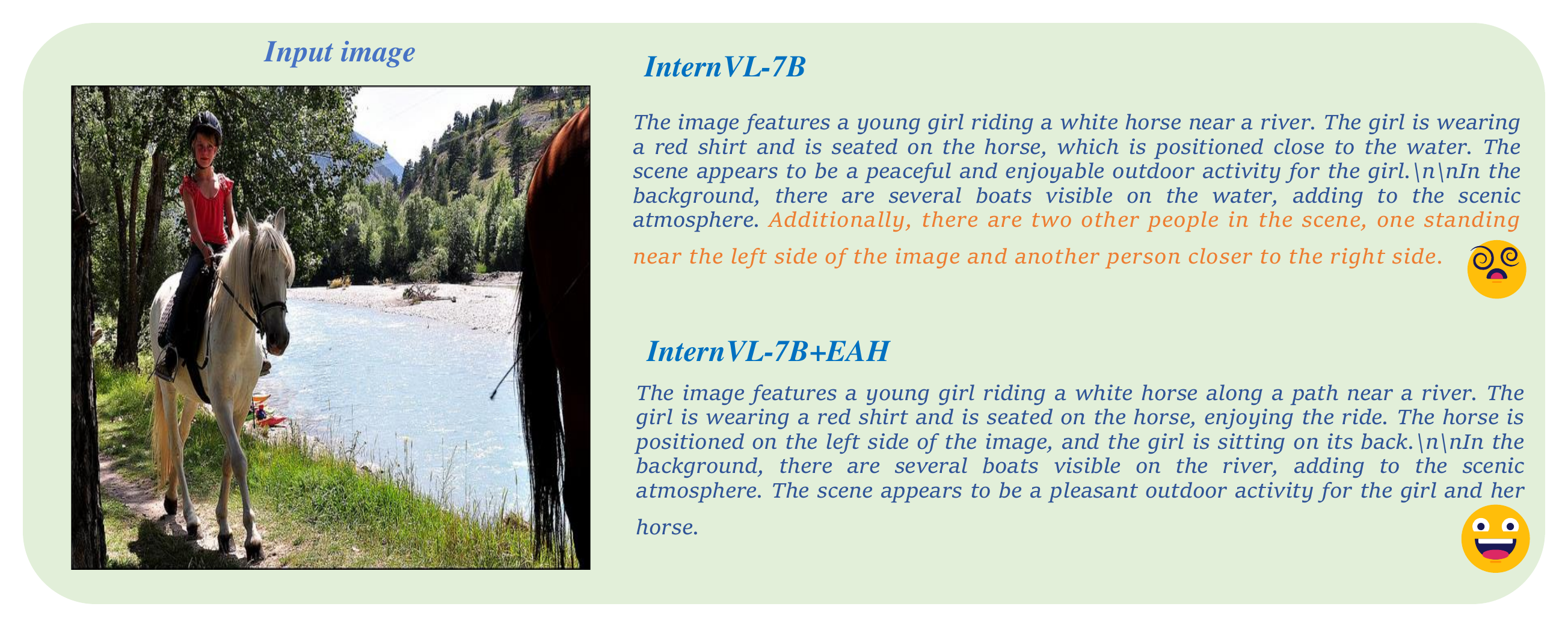}}
\caption{Results of Intern-VL with EAH, EAH can significantly reduce hallucinations while maintaining the original sentence length.}
\label{mgm+eah}
\end{figure*}

\begin{figure*}[h]
\centerline{\includegraphics[scale=0.37]{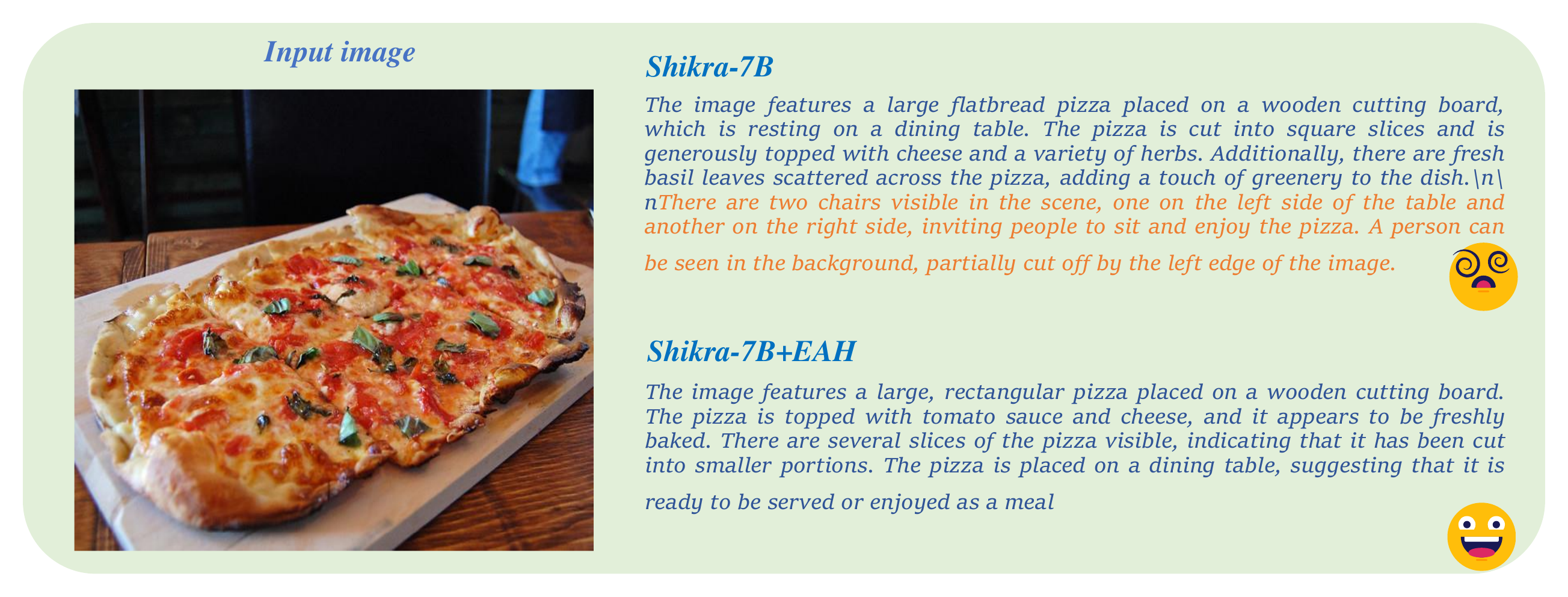}}
\caption{Results of Shikra with EAH, EAH can significantly reduce hallucinations while maintaining the original sentence length.}
\label{mgm+eah}
\end{figure*}

\end{document}